\RequirePackage{fix-cm}

\documentclass[10pt,journal,letterpaper,compsoc]{IEEEtran}

\usepackage{graphicx}
\usepackage{amsmath,amsfonts,amssymb}
\usepackage{subfigure} 
\usepackage{todonotes}
\graphicspath{{./images/}}
\usepackage{pgfplots}
\usepackage{tikz}
\usetikzlibrary{shapes.misc}
\usetikzlibrary{patterns}
\usepackage{placeins}
\usepackage{algorithm}
\usepackage{algorithmic}
\usepackage{url}
\newcommand{\bs}[1]{\ensuremath{\boldsymbol{#1}}}
\newcommand{\RNum}[1]{\uppercase\expandafter{\romannumeral #1\relax}}
\newcommand{\comment}[1]{\marginpar{\color{red}\sloppy\tiny #1}}

\renewcommand{\comment}[1]{}
\renewcommand{\color}[1]{}

\begin{document}

\title{Detecting Novel Processes with CANDIES -- An Holistic Novelty Detection Technique based on Probabilistic Models}


\author{Christian Gruhl         \and
        Bernhard Sick 
}


	\author{Christian Gruhl, Bernhard Sick
        \thanks{C. Gruhl and B. Sick are with the University of Kassel, Department of Electrical Engineering and Computer Science, Wilhelmshoeher Allee 73, 34121 Kassel, Germany (email: {cgruhl,bsick}@uni-kassel.de).}}

\date{Received: 15. April 2016}

\newcommand{\theApproach}{CANDIES}
\newcommand{\theApproachFull}{Combined Approach for Novelty Detection in Intelligent Embedded Systems}
\newcommand{\theApproachFirstUse}{\theApproachFull (\theApproach)~}

\maketitle
\comment{Name nochmals bzgl. Review angepasst}

\begin{abstract}
\color{blue}
In this article, we propose CANDIES (Combined Approach for Novelty Detection in Intelligent Embedded Systems), a new approach to novelty detection in technical systems. We assume that in a technical system several processes interact. If we observe these processes with sensors, we are able to model the observations (samples) with a probabilistic model, where, in an ideal case, the components of the parametric mixture density model we use, correspond to the processes in the real world. Eventually, at run-time, novel processes emerge in the technical systems such as in the case of an unpredictable failure. As a consequence, new kinds of samples are observed that require an adaptation of the model.
CANDIES relies on mixtures of Gaussians which can be used for classification purposes, too. New processes may emerge in regions of the models' input spaces where few samples were observed before (low-density regions) or in regions where already many samples were available (high-density regions). The latter case is more difficult, but most existing solutions focus on the former. 
Novelty detection in low- and high-density regions requires different detection strategies. With \theApproach, we introduce a new technique to detect novel processes in high-density regions by means of a fast online goodness-of-fit test. For detection in low-density regions we combine this approach with a 2SND (Two-Stage-Novelty-Detector) which we presented in preliminary work.
The properties of \theApproach~ are evaluated using artificial data and benchmark data from the field of intrusion detection in computer networks, where the task is to detect new kinds of attacks.

\end{abstract}

\section{Introduction}
\label{intro}

Today, so-called ``smart'' or ``intelligent'' technical systems are often equipped with abilities to act in real environments that are termed to be ``dynamic'' in the sense that their characteristics are time-variant (change over time). But typically, knowledge about the basic nature of these changes is built into these systems and it is assumed that only the time when these changes occur cannot be predicted. Future systems, however, have to evolve over time. Not all knowledge about any situations the system will face at run-time will be available at design-time. That is, the system has to detect and react on fundamental changes in \textit{time-variant environments}. As an example, we may consider technical systems that make observations of their environment with sensors and classify these observations (samples).
In a \textit{time-invariant environment}, there may be different causes for different kinds of observations (called ``processes'' in the following). The data are modeled (e.g., by means of probabilistic models such as Gaussian mixtures) and then a classifier is built (e.g., by gradually assigning components of the Gaussian mixture to classes). At run-time, the data model and the classifier are not adapted.

\begin{figure}[ht!]
    \vskip 0.2in
    \begin{center}
        \includegraphics[width=.75\columnwidth]{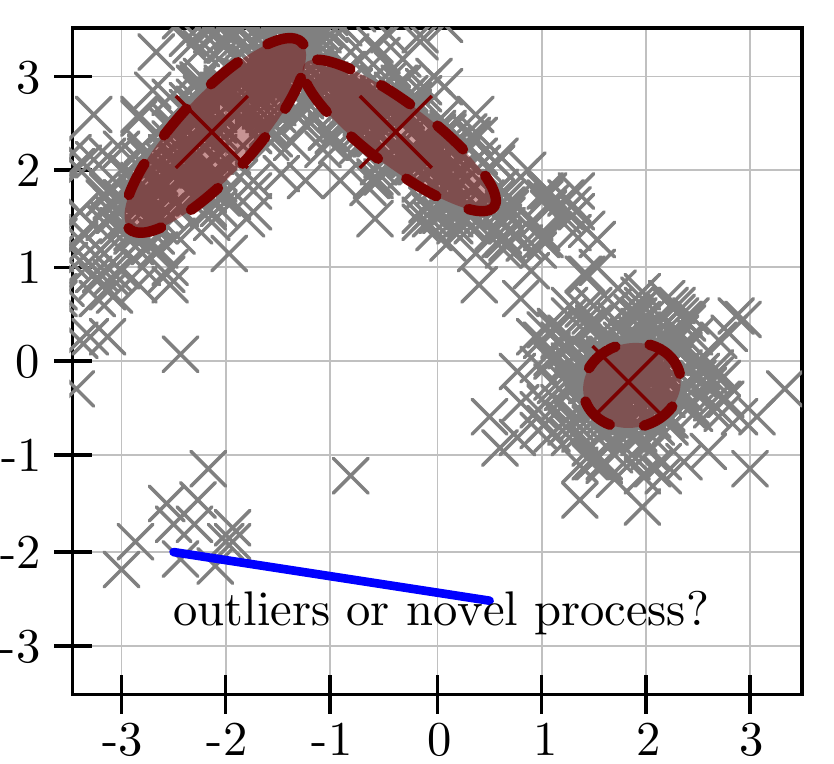}
        \caption{A situation with samples produced by three processes (represented by three components and marked in red) and a model of the situation with three components. Some observations appear in a \textit{low-density region} and are ``suspicious'', they may indicate that a novel process currently emerges.\label{fig:intro}}
    \end{center}
    \vskip -0.2in
\end{figure}

{\color{blue}
An example for such a system could be a machine, that produces various parts (cf. left part of Figure \ref{fig:example_scenario}). A \textit{process} in this hypothetical environment would be similar to the production of a certain part. If the system is monitored with multiple sensors (e.g., \textit{S1} and \textit{S2}), the resulting sensor signals span a two-dimensional \textit{input space} (as shown in the middle part of Figure \ref{fig:example_scenario}). Each value pair is called an \textit{observation} or \textit{sample}. With suitable machine learning techniques we can approximate the resulting distribution of \textit{samples}. On the right side of Figure \ref{fig:example_scenario} a Gaussian mixture model (GMM, cf. \ref{sec:gmm}) is used to approximate, and thus modeling, the sample distribution. Ideally, each component of the GMM describes a physical \textit{process} in the environment. Reasons to rely on GMM for this purpose is that arbitrary continuous densities can be approximated by GMM (with any desired precision, based on the number of components) and the generalized central limit theorem, which states that the sum of i.i.d. random samples tends to be normally distributed (assumed, that the variance is finite). 
{\color{red}In technical system this is frequently the case, since observed sensor values are often the outcome of various random parameters that influence each other.}
\begin{figure}
    \centering
    \includegraphics[width=.15\textwidth]{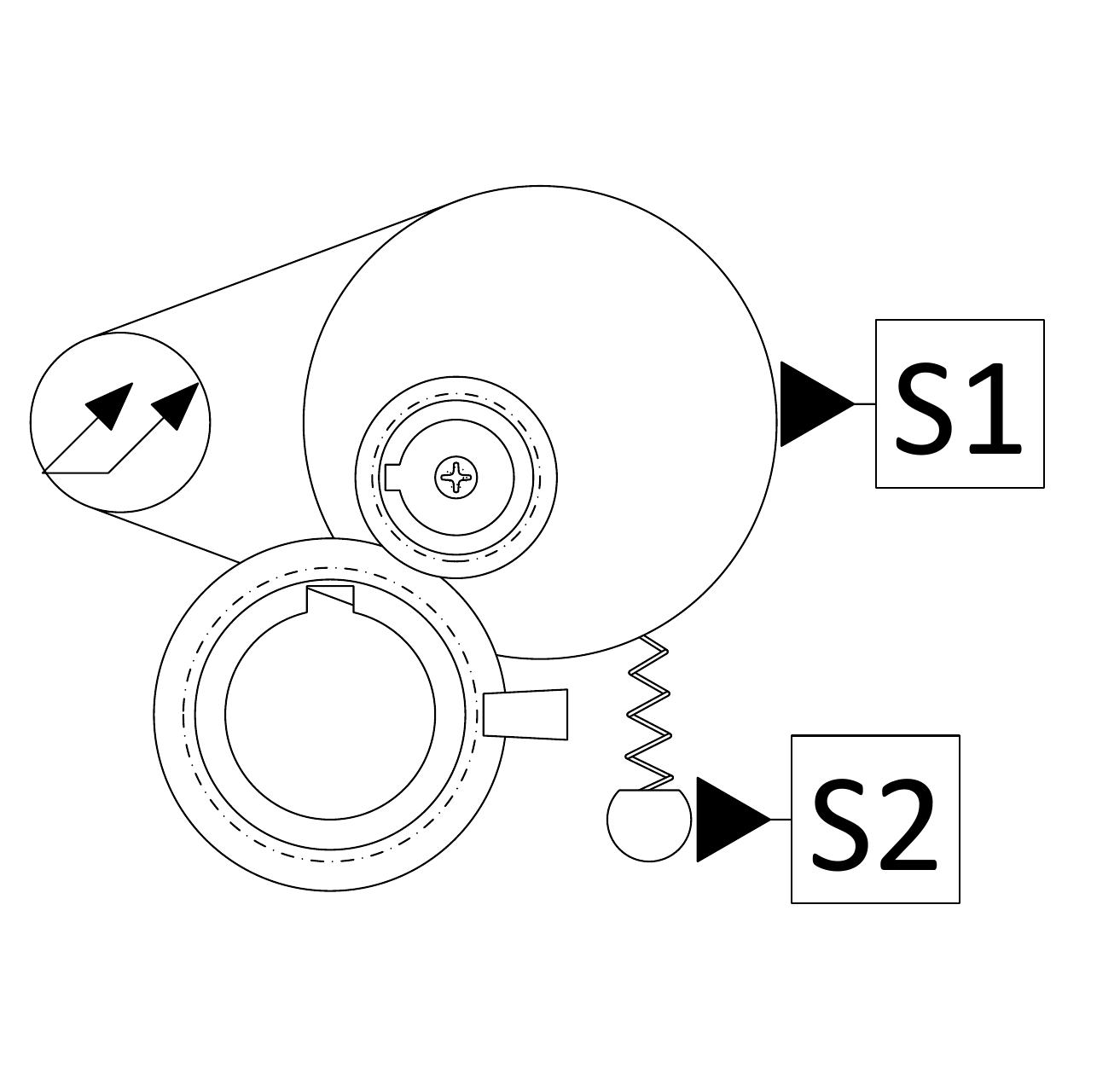}    
    \includegraphics[width=.15\textwidth]{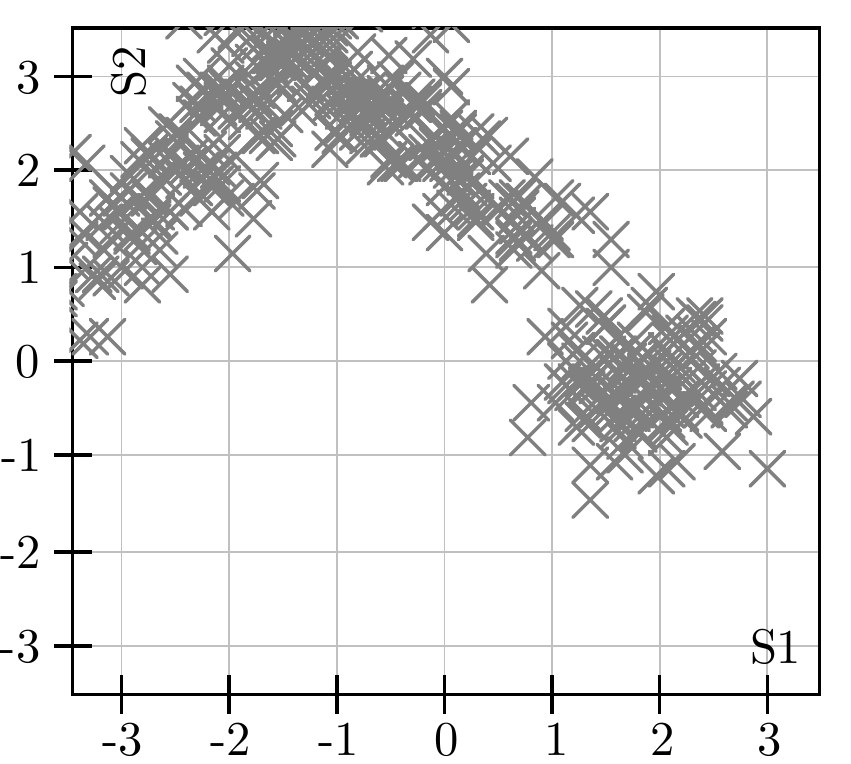}    
    \includegraphics[width=.15\textwidth]{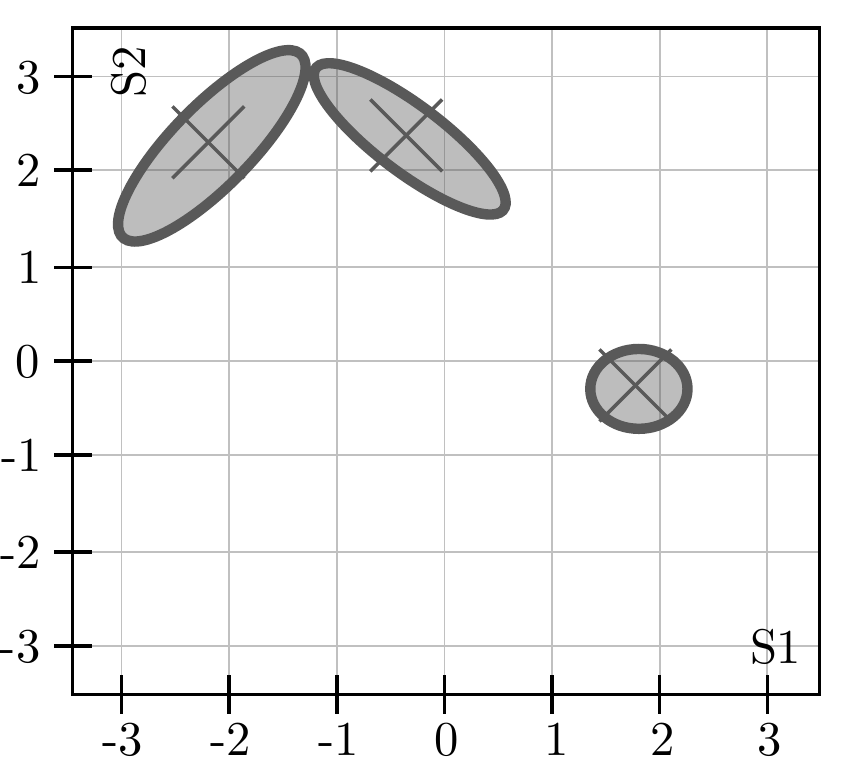}
    \caption{Hypothetical scenario of a monitored machine. On the left: abstract machine with monitoring sensors S1 and S2. In the middle: the two-dimensional \textit{input space} consisting of measured sensor signals from S1 and S2. In this case, the outcomes of three different processes are gathered in three clusters.  On the right: approximated density model, the ellipses are called \textit{components} and correspond to multivariate Gaussians that represent the physical \textit{processes}.\label{fig:example_scenario}}
\end{figure}
}
In a \textit{time-variant environment}, there may eventually be some conspicuous samples (cf.\ Fig.\ \ref{fig:intro}). Then (if the system is able to detect such a situation), some questions come up: Are these samples outliers of existing processes or not? If not, is there an anomaly in the observed environment or did a \textit{new} process emerge that was unknown at design-time? And how can we build \textit{novelty detection} methods to identify such \textit{novel} processes?\comment{Hier nochmal explizit Novelty Detection, def. sollte aus Kontext hervorgehen? Ist etwas im Widerspruch zu der Annahme "if the system is able to ..."} How can we adapt the data model to suit the changed environment and when (in order to find a trade-off between fast and accurate reaction)? And if there are new model components, to which class do we have to assign them?

\begin{figure}[ht!]
    \vskip 0.2in
    \begin{center}
        \includegraphics[width=.75\columnwidth]{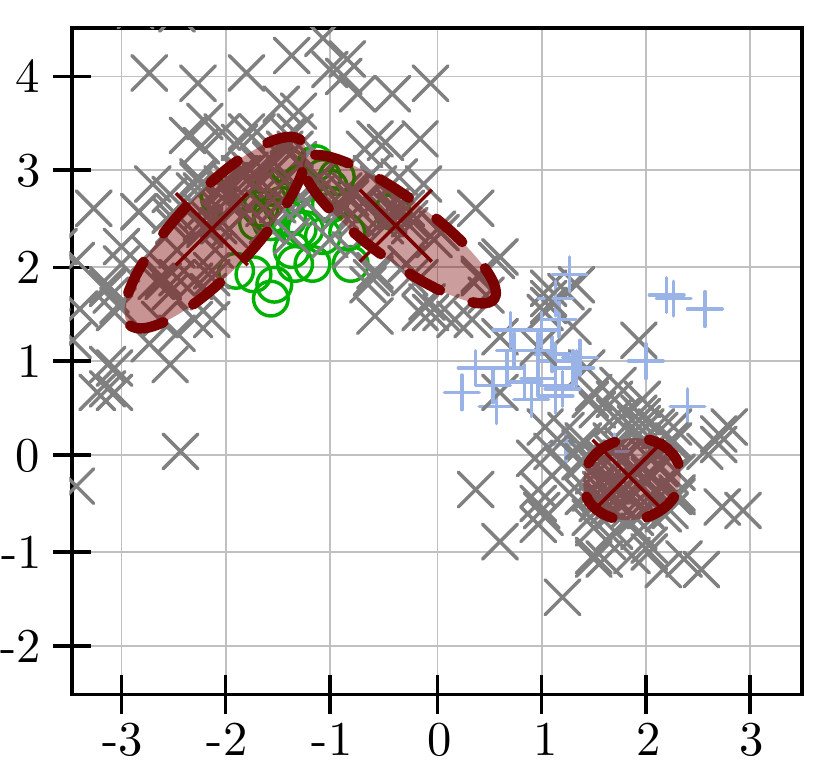}
        \caption{Same situation with samples produced by three processes (represented by three components and marked in red) and corresponding model. Some observations (green circles $\circ$ and blue crosses $+$) in the \textit{high-density region} covered by the model components, are the outcome of two not yet known processes and should be considered to be ``suspicious''.\label{fig:intro_hdr}}
    \end{center}
    \vskip -0.2in
\end{figure} 
{\color{blue}
With \theApproach~(\theApproachFull), some of the questions can be answered. 
One major challenge is the reliable detection of (possibly multiple) \textit{novel processes} in the complete input space.
Herby we assume that the input space is divided into two parts:
\begin{enumerate}
    \item \textit{High-density regions} (HDR): These are regions that are already covered by one or more components of the  mixture model (i.e.,~the support of the kernels, in our case Gaussians, is high). This implies that \textit{normal} observations are expected to appear in these regions and thus forming the \textit{normal model}. However, new processes might also emerge here (e.g., ``close'' to, or between existing components, or even totally overlapping these) and therefore change the characteristics of the approximated density. We assume that HDR can be considered to be spatially compact in the input space, and that they contain the majority of the overall density mass.
    \item \textit{Low-density regions} (LDR): These regions are distant from any component centers, resulting in a low support of the kernels. Thus, \textit{normal} data is not expected to be observed here and observations appearing here are considered to be \textit{suspicious}. In contrast to HDR we assume, that LDR are widely spread in the input space and that usually only a single LDR exists (i.e., not separated by HDR).
\end{enumerate}
The transition between HDR and LDR is not strictly defined and is application dependent.
Caused by their different characteristics, different problems are faced to detect \textit{novel processes}. Since LDR {\color{red}have a potentially infinite support}, the main difficulty is to efficiently find spatial relations (i.e.,~clusters) between \textit{suspicious observations}.
On the other hand, for HDR, two issues must be addressed: 1) which observations are assumed to be \textit{normal} (outcome of an already known and modeled process) and which are \textit{suspicious} (i.e.,~outcome of a \textit{novel process}, or anomalies). 2) When is a \textit{novel process} present.

In a preliminary article (see \cite{Gru15}), we presented 2\-SNDR, an approach to solve the \textit{novelty detection} problem sket\-ched above for situations, where novel processes start to ``generate'' data in LDR of a probabilistic knowledge model (based on Gaussian mixtures)\comment{Das geklammerte mit den Gaussians weg?}. Figure \ref{fig:intro} depicts such an exemplary scenario (where a \textit{novel process} is emerging in a LDR).
\comment{Das ist von der 'alten' Version übernommen.}To detect \textit{novelty} in HDR \theApproach~ relies on a new approach that is premised on statistical goodness-of-fit testing (i.e., measuring how well observed samples fit the assumed distribution), adjusted to suite Gaussian mixture models (GMM, cf. Section \ref{sec:gmm}) and online environments. Figure \ref{fig:intro_hdr} shows a different situation, where two \textit{novel processes} started to ``generate'' samples in a HDR, but are not yet represented in the current model.
}

Altogether, it is possible to address a specific kind of time-variance in the observed environment which is useful for many applications. We may imagine other kinds of situations where processes disappear (\textit{obsoleteness}) or change some basic parameters (\textit{concept shift} or \textit{concept drift}). Our current research addresses these situations as well.

The remainder of this article is organized as follows: Section \ref{sec:state-of-the-art} gives a broad overview of related work, including other common novelty detection techniques, related topics, and what are the distinctions to \theApproach.
Section  \ref{foundations} briefly summarizes methodical foundations essential for this article.
Preliminary to the technical in depth details, a simplified overview of the idea behind the proposed technique is given in Section \ref{sec:overview}.
The main body, introducing \theApproach~ in detail, is contained in Section \ref{sec:online}.
In Section \ref{sec:casestudy} a small case study based on the KDD Cup 99 Computer Intrusion data set is presented.
Finally, a conclusion and outlook to future work is given in Section \ref{sec:outlook}.


\section{Overview of CANDIES}
\label{sec:overview}

\comment{Hier, oder an den Anfang von Abschnitt \ref{sec:online}?}

With \theApproach~ we aim on three main goals: 1) Detecting clusters of \textit{suspicious} samples (i.e.,~those that differ notably from what is expected). 2) Detecting such clusters in the complete input space, that is, in LDR and in HDR.
3) using the discovered clusters to model new \textit{processes}.

The algorithms consists of multiple detectors for HDR and a single one for the LDR. It works (simplified) in the following manner. The foundation of the whole approach is a GMM, that provides a density estimate of the expected data. An advantage of GMM is that they can easily be extended to a classifier and that they belong to the family of generative models, thus additional structural information about the expected observations can be deduced (in contrast to discriminative classifiers, e.g., SVM).
At first a new sample $\bs{x'}$  located either in a HDR or in a LDR. Depending on the location it is marked as \textit{normal} (located in HDR) or \textit{suspicious} (located in LDR) (\textit{suspicious} is what comparable algorithms mark as \textit{novelties}). Depending on that decision either the LDR detector or one of the HDR detectors is responsible for handling the new sample. If the sample is marked as \textit{suspicious} the sample is stored in a ring buffer on which a nonparametric clustering is performed. If a cluster in the buffer reaches a certain size the detector will report the detection of a \textit{novel process}.
Otherwise, when the sample is regarded as \textit{normal}, it is used to update one of the HDR-Detectors (there is one HDR-Detector for each individual component of the GMM), the decision which detector is updated is made at random. The HDR-Detector works by testing how well the last $m$ samples are fitting the estimated Gaussian bell. This is done by using a $\chi^2$ test. If the $t$-value exceeds the critical value the detector reports the detection of a \textit{novel process}. 

\section{Related Work}
\label{sec:state-of-the-art}

The main task for a Novelty Detector is to distinguish if a previously unseen \textit{sample} belongs either to a \textit{normal} model or if it is \textit{different} in some way so that it does not belong to the \textit{normal} data and is therefore \textit{novel}.
Closely related to the topic are the fields of anomaly and outlier detection.
Over a decade ago it was sufficient to roughly group novelty detection approaches into two classes: either statistical (cf. \cite{novel1}) or neural network based (cf. \cite{novel2}). 

Most of the statistical approaches are relying on a model of the processed data.
Observations are identified as (potentially) \textit{novel} if they differ to much from what is expected, e.g.,~described by an appropriate model. Further, these approaches can be discerned based on the models they are using -- either parametric or nonparametric models.
Novelty detection techniques based on nonparametric density modeling are, for example, those using \textit{k}-nearest neighbors approaches or kernel density estimators, see \cite{Yeung2002} for a sample application in intrusion detection. Parametric models on the other side make assumptions about the distribution of the observed samples, e.g.~Gaussian mixture models.
In preliminary work \cite{Fis12} we detect \textit{novelty} based on a parametric Gaussian mixture model and a state variable which monitors how well the observations fit the model.
The approach is used for comparison to \theApproach~ and briefly presented in the case study in Section \ref{sec:casestudy}.

The second group comprises detection techniques that are based on neural networks, e.g.,~multi-layer perceptrons, radial basis function neural networks, \cite{Bis94,Bon98} but, according to Markou and Singh \cite{novel2}, also include methods based on support vector machines, e.g.~ One-Class SVM as described by Tax and Duin \cite{Tax02}.

Since the early 2000s, the topic draw much attention as objective of research and changed considerably (i.e. new ranges of applications or whole new techniques, due to advances in computing power).
Now, a more recent survey \cite{Pimentel2014} suggest five different categories to group novelty detection approaches:
\textbf{i)} probabilistic, \textbf{ii)} distance-based, \textbf{iii)} reconstruction-based, \textbf{iv)} domain-based, and \textbf{v)} information theoretical.

The first category covers a large part of the approaches that where previously affiliated with the \textit{statistical} group. Typically these techniques are build upon a parametric density estimation of training data as a model. Frequently used are mixtures of Gaussians (\cite{Fis12} \cite{Ilonen2006,Zorriassatine2005}, for instance). Novelty is usually detected if samples are observed in \textit{low-density-regions} (i.e., the density for the observed sample is below a selected threshold). Several method to define a threshold are based on Extreme Value Theory (EVT, cf. \cite{Clifton2011,Hazan2012,Roberts1999}). The idea in EVT is to estimate the distribution of extreme values (i.e., maximum or minimum for legit samples) for a given density model and a given sample size. Then, samples that exceed the expected maximum or surpass the expected minimum are identified as \textit{novel}.
Recently Extreme Learning Machines with decision making depending on EVT where proposed by Al-Behadili \textit{et al.} \cite{Al-Behadili2015} to implement incremental semi-supervised learning based on novelty detection.	
Thus, probabilistic approaches are not limited to generative models, cf. \cite{Clifton2014}, for example, where Support Vector Machines are used for detection and resulting novelty values calibrated in order to be interpreted as class-conditional probabilities.

To the second category belong approaches that are based on distances. Popular representatives of this category are approaches based on $k$-nearest-neighbors ($k$nn). E.g.,~ \cite{Breunig2000} or \cite{Hautamaki2004,Papadimitriou2003}, where the latter use the density of a $k$-neighborhood (i.e.,~a radius required to enclose $k$ neighbors) to identify \textit{novel} samples. A sample is \textit{novel} if its neighborhood density is considerably lower than the density of its neighbors.
Clustering based approaches refer also to category \textbf{ii)}. Typically, \textit{normal} samples are aggregated to form clusters, \textit{novelty} is then determined by the minimal distance of an unseen sample to any centroid (e.g.~\cite{Spinosa2009,Wang2009}).
It is questionable whether category \textbf{i)} and \textbf{ii)} are sharply differentiable. Gaussian Mixture Models for example, consists of multiple location invariant kernels and the density is finally greatly dependent on the applied distance measure.

Our new \theApproach~ approach does not fit into a single category but is a hybrid in the sense that it belongs to the first two categories: probabilistic and distance-based. For a detailed summary of the remaining categories \textbf{iii)}, \textbf{iv)}, and \textbf{v)} cf. \cite{Pimentel2014}.

However, most of the introduced paradigms are designed to spot only single samples as \textit{novelties} and do not relate those samples to one another. Thus, potential new knowledge (structural information in form of a cluster, that is evidence of a \textit{novel physical process}) is unexploited and discarded.
In some common applications such as medical condition monitoring \cite{Clifton2011a,Roberts2000,Tarassenko1995} or machinery monitoring \cite{Pontoppidan2003}, this is not a real drawback, since \textit{anomalies} might arose everywhere in the input space and are very specifically stuck to a concrete application (i.e., monitoring a specific patient or a specific engine). But in other fields, such as network intrusion detection, this discovered \textit{knowledge} has great potential to be used to detect future attacks.

The contributions of this article are:
\begin{enumerate}
\item \theApproach~ is trimmed to detect \textit{novel} processes (clusters of \textit{suspicious} observations, cf. \textit{knowledge}) in such a way, that the process can easily be integrated as new component into the existing GMM. \comment{Added component.}
This leads to the result, that \textit{learning} does not only happen in a isolated training phase, off-line at design-time, but it is also conducted at run-time \cite{Haehner2015}.
\item Novelty is not only detected in \textit{low-density} regions (where \textit{normal} observations are not likely to appear), but also in \textit{high-density} regions, i.e., where \textit{normal} observations are expected.
\end{enumerate}

\section{Methodical Foundations}
\label{foundations}
For the purpose of a self-contained article, we briefly recap the most important techniques used to implement our approach. This includes an overview of Gaussian mixtures, a method to extend those to a
classifier, a short introduction to nonparametric density estimation as foundation for cluster analysis, and a brief description of statistical goodness-of-fit testing.

\subsection{Gaussian Mixtures}

\label{sec:gmm}

One frequently used approach to generative modeling is the Gaussian mixture model (GMM). That is, a superposition of multiple multivariate normal distributions (denoted as $\mathcal{N}$ and commonly referred to as Gaussian) and a mixing coefficients $\pi_j$ (Eq.\ \eqref{eq:gmm}).
Each Gaussian is called a \textit{component} and has its own set of parameters which are the mean vector $\boldsymbol{\mu}_j \in \mathbb{R}^D$ and a covariance matrix $\boldsymbol{\Sigma}_j$ (with $dim(\boldsymbol{\Sigma}_j)=D\times D$) that describes its shape.
The mixing coefficients $\pi_j$ (with constraints $\sum_{j=1}^{J}\pi_j=1$, $\pi_j\in\mathbb{R}^+$) ensure that the resulting $p(\mathbf{x})$ ($\mathbf{x} \in \mathbb{R}^D$ is the random variable) still fulfills the requirements for a density function. They may also be interpreted as priors for each component (i.e,~the probability that an unobserved sample is generated by the corresponding component). Altogether, we get:
\begin{equation}
\label{eq:gmm}
p(\mathbf{x}) = \sum_{j=1}^{J} \pi_j \cdot \mathcal{N}(\mathbf{x}|\boldsymbol{\mu}_j,\boldsymbol{\Sigma}_j).
\end{equation} 
\begin{figure}[ht]
    \vskip 0.2in
    \begin{center}
        \subfigure[Training set with classes green circle $\circ$ and blue cross \textit{+}. The resulting GMM is trained in an unsupervised manner and models the density with three components.\label{fig:gmm}]{\includegraphics[width=.475\columnwidth]{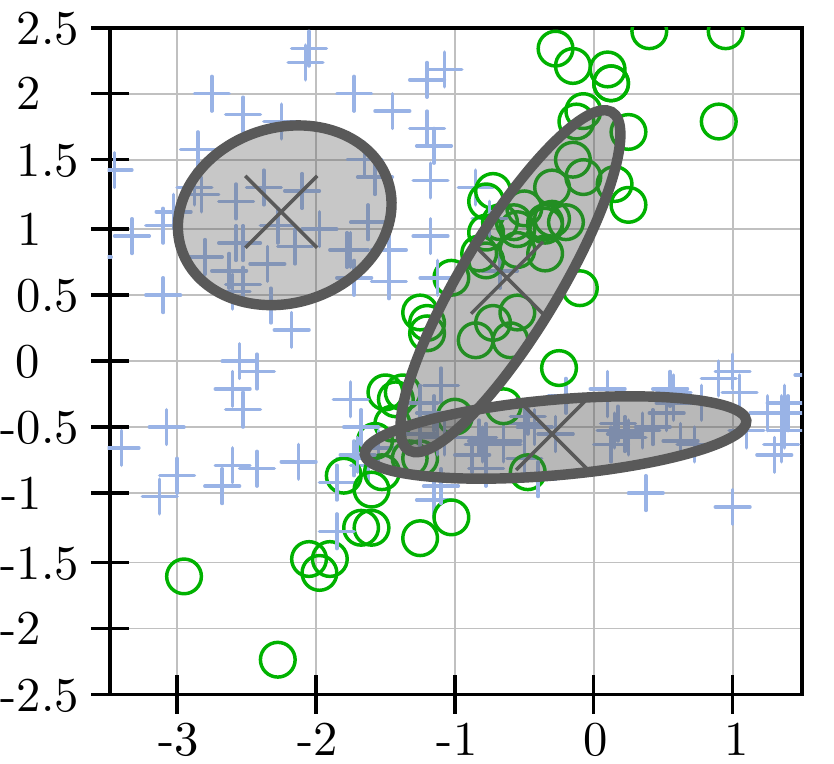}}
        \hspace{1mm}	
        \subfigure[Model with class conclusions extended to a classifier. The thick black line is the decision boundary that devides the input space into decision regions.
        \label{fig:cmm}]{\includegraphics[width=.475\columnwidth]{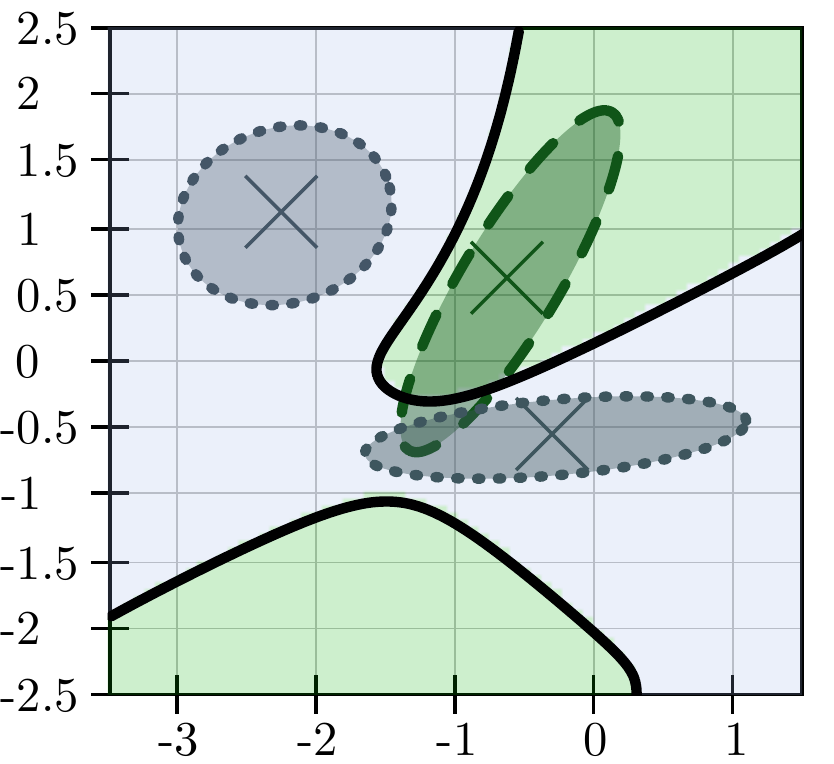}}	
        
        \caption{Each $\times$ denotes the center $\boldsymbol{\mu}_j$ of the j-th component while each ellipse represents the shape which is defined by the j-th covariance matrix $\boldsymbol{\Sigma}_j$. The distance between $\boldsymbol{\mu}_j$ to the associated ellipse, which is a constant density surface, corresponds to a Mahalanobis distance of 1.}
        \label{icml-historical}
    \end{center}
    \vskip -0.2in
\end{figure} 

An ordinary GMM models only the density of an associated training set and can
be trained in an unsupervised manner (i.e.,~labels are not required).
Since the sufficient statistics for the components cannot be computed in closed form, we pursue this goal with an expectation-maximization (EM) like approach that uses 2nd order (or hyper-) distributions and is heavily based on variational Bayesian inference (VI). An extensive introduction to VI is given by \cite{Bis06}.
For clarification, a trained GMM for a two-dimensional data set is shown in Figure \ref{fig:gmm}.

Relying on VI gives rise to two advantages: (1) prior knowledge about the data can be included, which is especially valuable in real-world applications,
and (2) multiple GMM can be \textit{fused} into one model as described in \cite{Fis14}.
The final GMM is obtained from the expectations (a point estimate from the second-order distributions) of the hyper-distributions after the VI training finishes.

Since we assume a certain functional form of the underlying distribution and estimate its parameters, GMM are parametric density models.

\subsection{Classification Paradigm}
\label{sec:cmm}

To derive a classifier $h(\mathbf{x})$ from the trained density mo\-del $p(\mathbf{x})$, we estimate the class posteriors $p(c|\mathbf{x})$ in a second, supervised (i.e.,~with respect to class labels) iteration. 
The classification of a given sample $\mathbf{x}$ is then done, as shown in Eq.\ \eqref{eq:cmm:map} by selecting the maximum a-posteriori (MAP) of the class probabilities:
\begin{equation}
\label{eq:cmm:map}
h(\mathbf{x}) = \underset{c}{\operatorname{argmax}}\left\{p(c|\mathbf{x})\right\},
\end{equation}
with 
\begin{equation}
\label{eq:cmm:post}
p(c|\mathbf{x}) = \sum_{j=1}^{J} p(c|j)\cdot p(j|\mathbf{x}) = \sum_{j=1}^{J} \xi_{j,c}\cdot \gamma_{\mathbf{x},j},
\end{equation}
where
\begin{align}
\label{eq:cmm:resp}
\gamma_{\mathbf{x},j} &=  \frac{\pi_j\mathcal{N}(\mathbf{x}|\boldsymbol{\mu}_j,\boldsymbol{\Sigma}_j) }{\sum_{j'=1}^{J}\pi_{j'}\mathcal{N}(\mathbf{x}|\boldsymbol{\mu}_{j'},\boldsymbol{\Sigma}_{j'})},\\
\label{eq:cmm:classpost}
\xi_{j,c} &= \frac{1}{N_j} \sum_{\mathbf{x}_n \in \mathbf{X}_c} \gamma_{\mathbf{x}_n,j}.
\end{align}

Eq.\ \eqref{eq:cmm:resp} shows the \textit{responsibilities} $\gamma_{\mathbf{x},j}$ which are the probability that a given sample $\mathbf{x}$ was generated by the \textit{j-th} component. For each component $j$ and class $c$ the conclusion is determined by Eq.\ \eqref{eq:cmm:classpost}, which is the fraction of all responsibilities for samples $\mathbf{x}_n \in \mathbf{X}_c$ that are labeled with class $c$ and the effective number of samples (denoted as $N_j = \sum_{n=1}^{N} \gamma_{\mathbf{x}_n,j}$) belonging to the \textit{j-th} component ($\mathbf{X}$ is the overall set of labeled samples, $\mathbf{X}_c$ the subset of $\mathbf{X}$ associated with class $c$).

Finally, the class posteriors $p(c|\mathbf{x})$ given in Eq.\ \eqref{eq:cmm:post} are a composition of the responsibilities $\gamma_{\mathbf{x},j}$ and the class conclusions $\xi_{j,c}$. The resulting decision boundary, which describes the classifier for the previously estimated density model, is shown in Figure \ref{fig:cmm}.

\subsection{Density Based Clustering}

Rather than assuming a specific functional form such as parametric methods, nonparametric techniques provide a point estimate for the density $p(\mathbf{x})$ at a given point $\mathbf{x}$.
One well-known nonparametric method is the Parzen window (or kernel) density estimator, here with Gaussian kernel:
\begin{equation}
\label{eq:parzen:estimator}
p(\mathbf{x}) = \frac{1}{N\cdot h^D} \sum_{n=1}^{N} k\left(\frac{\mathbf{x} - \mathbf{x}_n}{h}\right).
\end{equation}
It is the sum of a finite set of $N$ samples $\mathbf{x}_n$ of an underlying training set to which an appropriate \textit{kernel function} is applied to. The kernel is placed at the point $\mathbf{x}$ where the density should be estimated. The parameter $h$ is a \textit{smoothing} factor that controls how \textit{smooth} the estimation is while $D$ is the number of dimensions. Closely related to the Parzen window are histograms (cf. \cite{Bis06}).

The DBSCAN clustering algorithm (cf. \cite{Est96}) uses a density estimation that is quite similar to a Parzen window estimator. Based on the density at each sample the algorithm decides whether a sample belongs to, lies at the edge, or is outside a cluster (in that case it is considered as \textit{noise}). To do so, the kernel in Eq.\ \eqref{eq:dbscan:kernel}:
\begin{align}
k(\mathbf{x}) &= \begin{cases}
1, \hspace{5mm} \mathrm{if} \hspace{2mm} dist(\mathbf{x},\mathbf{0}) \leq \epsilon \\
0, \hspace{5mm} \mathrm{otherwise}
\end{cases} \label{eq:dbscan:kernel}
\end{align}
is used which forms an $D$-dimensional sphere around the point $\mathbf{x}$ 
with radius $\epsilon$. Typically, $dist$ is realized with an Euclidean metric. If a sample is part of a cluster, all samples inside the sphere are also assigned to the same cluster.
The advantage of this approach is that clusters of arbitrary shapes can be identified.

\subsection{Statistical Goodness-of-Fit Tests}
\label{sec:found:chi2}

\begin{figure}
    \centering
    \includegraphics[width=.75\columnwidth]{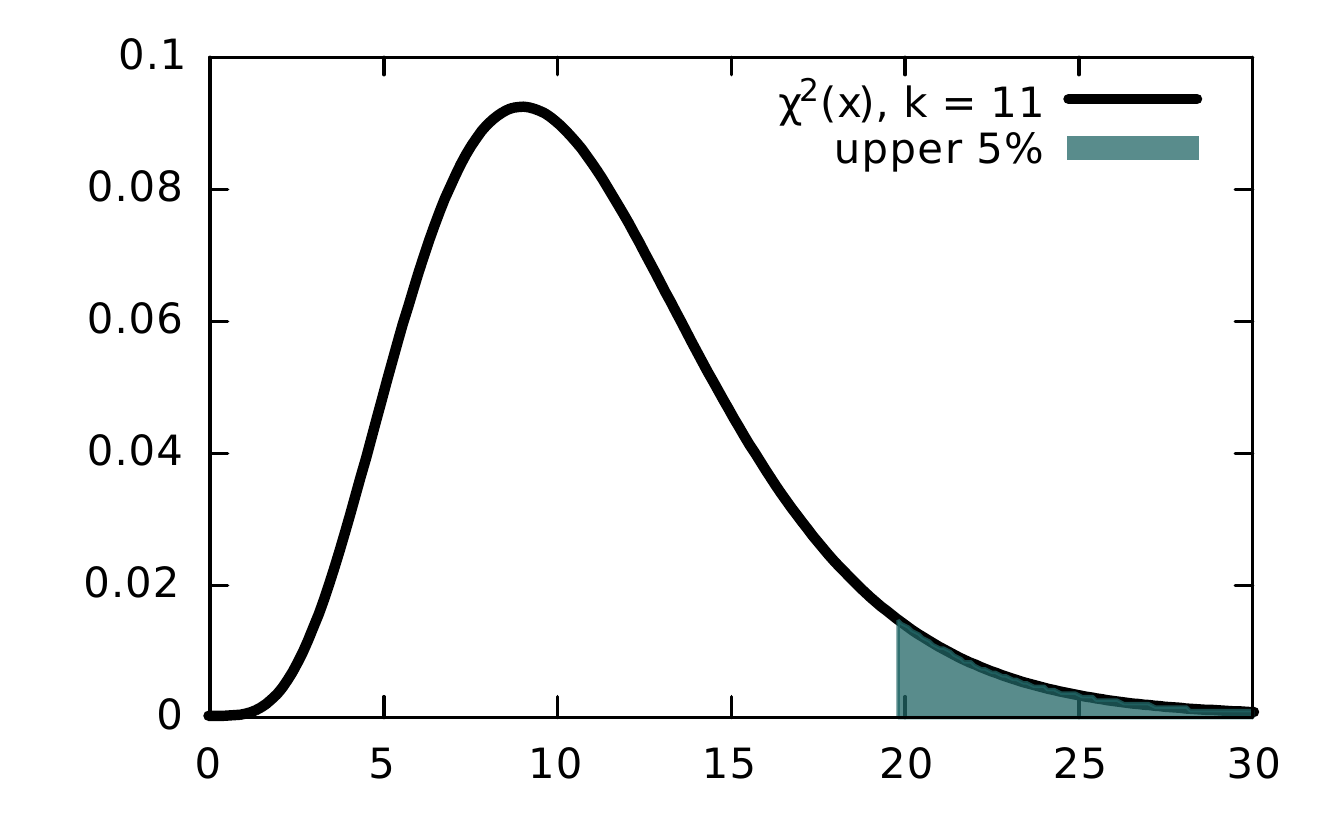} 
    \caption{Distribution of $t$-values for $\chi^2$ test with $11$ degrees of freedom. The marked region at the right is the rejection area for a significance level of $\alpha=5\%$. The beginning of the region is equal to the critical value $\chi^{2,upper}_{5\%,11}$. \label{fig:chi2test}}
\end{figure}

To validate whether an observed sample matches a hypothesized distribution or not, goodness-of-fit tests can be applied. A fast and reliable method is Pearson's chi-squared ($\chi^2$) test \cite{Pearson1900}.
The test compares observed frequencies from mutually exclusive events (finite set of possible outcomes/values of a discrete random variable)\comment{Ist damit Event klar?} against expected theoretical frequencies (obtained from a suitable fitted distribution) of these events.
The test statistic (or $t$-value) is calculated by:
\begin{align}
    t &= \sum_{i}^{k} \frac{(x_i - e_i)^2}{e_i} \label{eq:gof_tval}
\end{align}
where $x_i$ is the observed event frequency of event $i$ and $k$ is the total number of different events.
The expected frequencies of events $i$ are given by $e_i$:
\begin{align}
    e_i &= P_{\mathrm{\textbf{fit}}}(i|\Theta) \label{eq:gof_eval}.
\end{align}    
where $P_{\mathrm{\textbf{fit}}}$ is the fitted distribution. The test aggregates the squared deviations between observed and expected frequencies and weights them by the expected frequency. This leads to stronger penalties when only small frequencies are expected.
To accept or reject the \textit{null}-hypothesis that the sample is drawn from the hypothesized distribution the $t$-value must be less than the critical value:
\begin{align}
    \chi^{2,\mathrm{\textbf{upper}}}_{\alpha,k} &= F^{-1}_{\chi^2_{\nu}}(1 - \alpha) \label{eq:criticalValue}
\end{align}
The critical value is calculated by evaluating the inverse cumulative density function $F^{-1}$ of the $\chi^2$ distribution with $\nu$ degrees of freedom at point $1-\alpha$. Where $\alpha$ is the \textit{significance level} which implies that the error rate for type \RNum{1} errors is at most $\alpha$ (often $\alpha = 5\%$ or $1\%$).
The degrees of freedom $\nu$ are given by the number of events minus the number $p$ of covariate parameters $\Theta$ of the fitted density $P_{\mathrm{\textbf{fit}}}$ (e.g.,~$p=2$ for univariate Gaussian with $\Theta = (\mu, \sigma)$). Figure \ref{fig:chi2test} emphasizes the relation between the $\chi^2$ distribution of $t$-values and the critical value $\chi^{2,upper}$ for a significance level of $\alpha=5\%$.

\section{Online Novelty Detection}
\label{sec:online}

\comment{An Terminologie angepasst HDR, LDR}
This section is split into three parts: the first part is based on our previous work \cite{Gru15} and discusses novelty detection and reaction in LDR with \textbf{2SND}. The second part treats novelty detection in HDR with online capable \textbf{$\chi^2$} goodness-of-fit tests. The last part then introduces \textbf{CANDIES} a detector which is able to detect novelties in the whole input space by combining both previously mentioned techniques.
All techniques share the property to be applicable to online environments (i.e.,~soft real-time).
\subsection{Novelty Detection in Low-Density Regions}
\label{sec:2snd}

To detect \textit{novel processes} in sparse LDR we developed the \textbf{2} \textbf{S}tage \textbf{N}ovelty \textbf{D}etection (2SND) algorithm.
The algorithm works on top of an existing GMM or CMM (as described in Section \ref{sec:cmm}) and extends it with novelty detection capabilities. Further, with 2SND it is possible to update the underlying GMM/CMM and to enhance them by including components that model the detected \textit{novel processes}.

\comment{PROPAGATE not longer contained as algorithm}
{\color{red}
The algorithm itself consists of two procedures: a main procedure 2SND (Alg. \ref{alg:salmiac}) and an auxiliary procedure
PROPAGATE, that propagates the cluster id to all affiliated samples using a modified breadth-first search.
}

\begin{figure}[ht]
	\vskip 0.2in
	\begin{center}
		\subfigure[A normal distribution with mean $\mu=0$ and variance $\sigma^2=1$. The darker green area shows the region where $66\%$ of the probability mass  is located. The combination of both areas corresponds to a mass of $90\%$. Since the squared distances are $\chi^2_1$ distributed, the radii of the areas are equal to the
		root of the quantile function $F_{{\chi}_{D}^2}^{-1}$ of the $\chi^2_D$ distribution 
		which is $\sqrt{\rho}\approx0.95$ for the darker green area (blue line) and  $\sqrt{\rho}\approx1.64$ for
		the combined area (green line).The marked areas are identical to the respective high-density regions. \label{fig:chi2:1d}]{\includegraphics[width=.96\columnwidth]{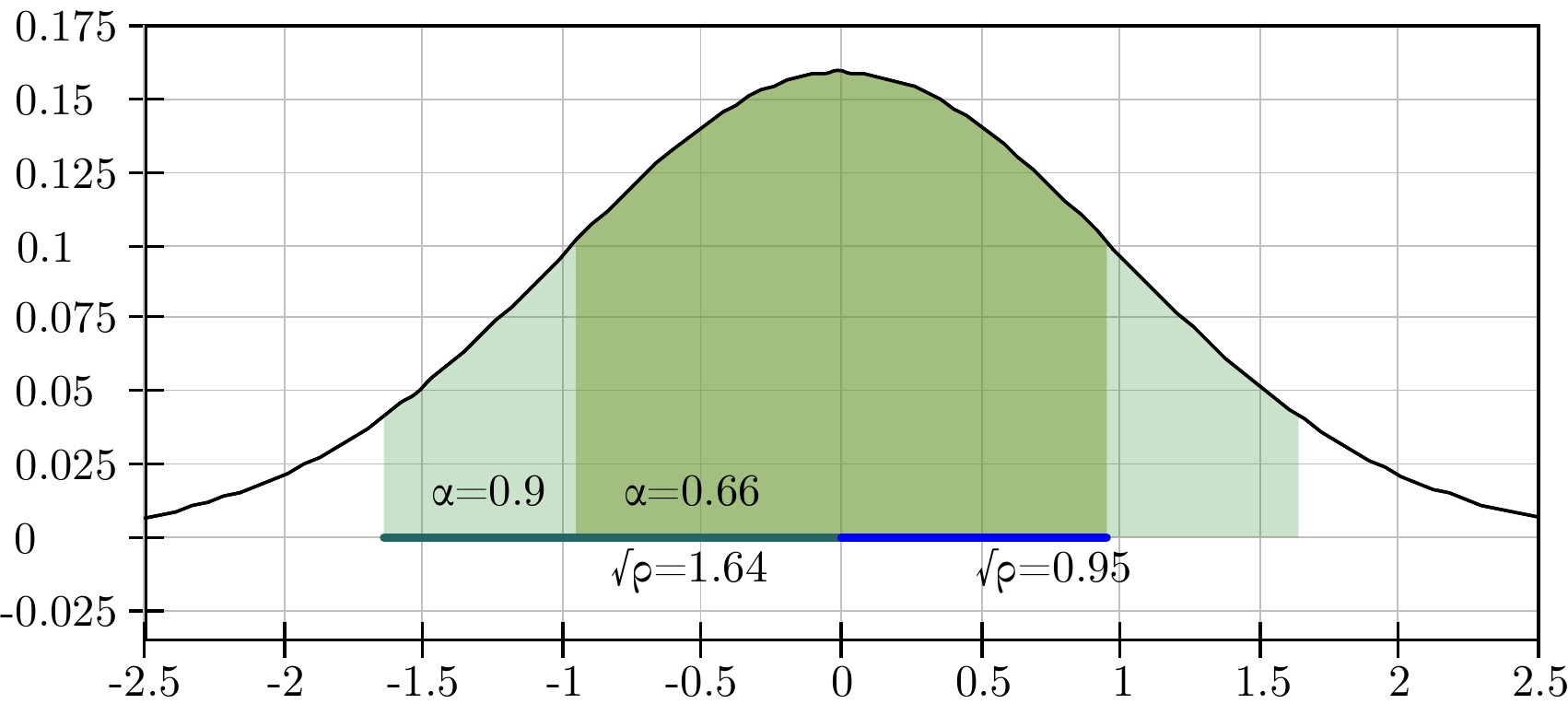}}
		\subfigure[Bivariate Gaussian with $90\%$ $\alpha$-region and maximum Mahalanobis distance of $\sqrt{\rho}=2.15$.\label{fig:chi2:2d85}]{\includegraphics[width=.475\columnwidth]{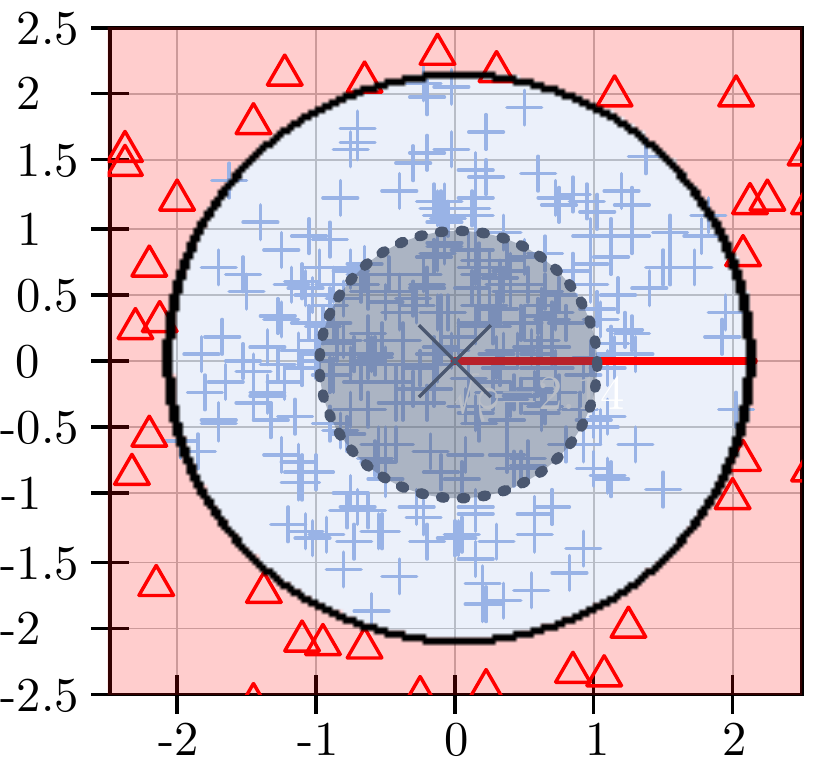}}	
		\hspace{1mm}
		\subfigure[Bivariate Gaussian with $66\%$ $\alpha$-region and maximum Mahalanobis distance of $\sqrt{\rho}=1.47$.\label{fig:chi2:2d50}]{\includegraphics[width=.475\columnwidth]{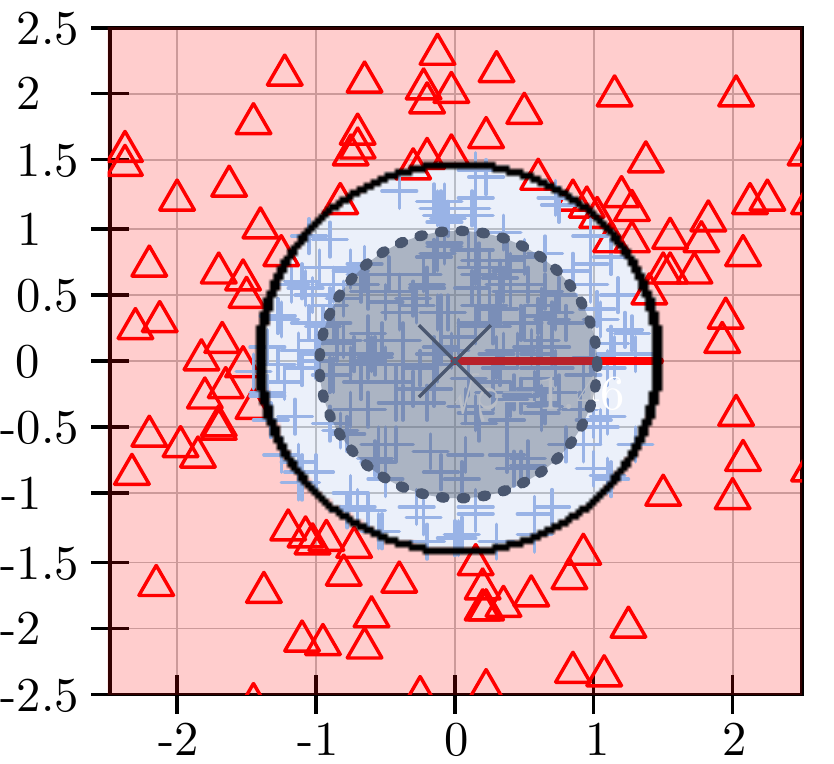}}	
		\caption{Relation between normal distribution and $\chi^2_D$ distributed distances. The dashed ellipses in \subref{fig:chi2:2d85} and \subref{fig:chi2:2d50} are level curves with a Mahalanobis distance of 1 while the black ellipses have a distance of $\sqrt{\rho}=\sqrt{F_{{\chi}_{2}^2}^{-1}(\alpha)}$ to their center. Samples displayed as red triangles $\vartriangle$ are \textit{suspicious} (potentially \textit{novel}), while samples depicted as blue circles $\circ$ are \textit{not suspicious}. High-density regions (HDR) are colored in blue, low-density regions (LDR) in red.}
		\label{fig:chi2}
	\end{center}
	\vskip -0.2in
\end{figure}

To detect \textit{novel} processes, we propose a two-stage approach which identifies \textit{suspicious} samples in the first stage and novel processes in the latter.
Each assessed sample is individually tested how well it suits the current model by determining whether it resides in a high- (HDR) or low-density region (LDR). \comment{nur abkürzungen?}
This is done by exploiting the fact that the squared Mahalanobis distances between samples from a Gaussian $j$ to its mean $\boldsymbol{\mu}_j$:
\begin{equation}
\label{eq:mhd}
\Delta^2_j(\mathbf{x}) = (\mathbf{x}-\boldsymbol{\mu}_j)^\mathrm{T}\boldsymbol{\Sigma}^{-1}_j(\mathbf{x}-\boldsymbol{\mu}_j)
\end{equation}
are $\chi^2_D$-distributed, where $\boldsymbol{\Sigma}^{-1}_j$ is the inverted covariance (or the precision) matrix. 
With the quantile function $F_{{\chi}_{D}^2}^{-1}$ of the $\chi^2_D$ distribution, we can determine a squared Mahalanobis distance $\rho = F_{{\chi}_{D}^2}^{-1}(\alpha)$ such
that a fraction $\alpha$ of samples (which belong to the Gaussian) have a smaller squared Mahalanobis distance to the mean as $\rho$. Figure \ref{fig:chi2} depicts the relationship for one- and two-dimensional Gaussians.

\comment{nur abkürzungen? HDR/LDR?}

Separating the input space into HDR and LDR simplifies the model considerably. Legitimate samples are assumed to appear in the dense regions while samples in the low-density regions are less likely to be observed. To detect \textit{novel processes} in HDR additional detectors are required (cf. Section \ref{sec:onlineGoF}), since 2SND focuses only on LDR.
By selecting $\alpha$ we specify how much of the total probability mass is covered by HDR, thus defining the transition between HDR and LDR. Samples with a Mahalanobis distance of $\Delta_j(\bs{x}) \leq \rho$ to at least one of the component centers $\bs{\mu_j}$ are located within a HDR and therefore seen as \textit{not suspicious}. Samples with a higher distance $\Delta_j(\bs{x}) > \rho$  to all centers are regarded as being \textit{suspicious}. 
This complies to the first stage of our novelty detection.

Figures \ref{fig:chi2:2d85} and \subref{fig:chi2:2d50} are illustrate an exemplary situation for a GMM with a single component and different values of $\alpha$. Observations inside the $\alpha$-region (which is equal to a HDR) are depicted by circles $\circ$, while suspicious samples (located in a LDR) are shown as triangles $\vartriangle$. The first stage is implemented in the first part of the procedure  2SND given in Alg. \ref{alg:salmiac}.


The second stage utilizes a density based clustering approach. Each sample $\mathbf{x'}$ that is identified as being suspicious is cached in a circular buffer $\mathcal{B}$ with size $\tilde{b}$.
Based on the distance to the nearest neighbor of $\mathbf{x'}$ in $\mathcal{B}$, the algorithm decides if $\mathbf{x'}$ belongs to an already existing cluster.
This behavior depends on the kernel given in Eq.\ \eqref{eq:dbscan:kernel} with $\epsilon$ being the maximum distance between a sample $\mathbf{x'}$ and its nearest neighbor and it is implemented in the second part of the 2SND procedure, given in Alg.\ \ref{alg:salmiac}.
If the sample is associated with a cluster $C$, the cluster is extended to include all $\epsilon$-neighbors (i. e.,~all buffered samples with a distance $dist(\mathbf{x'},\mathbf{x_\mathcal{B}}) \leq \epsilon$). This is achieved with the procedure PROPAGATE, which is basically a breadth-first search with constraints.
In fact, expanding a cluster can lead to a merger of multiple clusters and, thus, create a much larger cluster.

A \textit{novel} process is detected as soon as a cluster $C$ fulfills the adaptation criterion $|C| \geq minPts$ which corresponds to the number of samples that are associated with the cluster $C$.
In Section \ref{sec:measures} a measure is proposed to represent the current amount of \textit{novelty} in LDR in human readable form.

\begin{algorithm}[th]
	\caption{2SND}
	\label{alg:salmiac}
	\newcommand{\CGRETURN}{\STATE\textbf{return} }
	\newcommand{\CGEMPTY}{\STATE{\vspace{2.5mm}}}
	\begin{algorithmic}
		\STATE {\bfseries Input:} sample $\mathbf{x'}$, parameters $\alpha$,$\epsilon$,$minPts$,$\tilde{b}$
		\STATE {\bfseries Global:} model $\mathcal{M}$, buffer $\mathcal{B}$
		\STATE Initialize $\rho =  F_{{\chi}_{D}^2}^{-1}(\alpha)$.
		
		\CGEMPTY
		\COMMENT{1st stage -- detection of suspicious samples}
		\FORALL{components $j$ in $\mathcal{M}$}
		\IF{$\Delta^2_j(\mathbf{x'}) \leq \rho$}
		\item \COMMENT{The observation is not suspicious}
		\CGRETURN classification of $\mathbf{x'}$ based on $\mathcal{M}$.
		\ENDIF
		\ENDFOR
		
		\IF{$|\mathcal{B}| = \tilde{b}$}
		\STATE Remove oldest sample from buffer $\mathcal{B}$.
		\ENDIF
		\STATE Add \textit{suspicious} sample $\mathbf{x'}$ to buffer $\mathcal{B}$.
		\CGEMPTY
		\COMMENT{2nd stage -- detection of novel processes}
		
		\STATE Find nearest neighbor $\mathbf{nn}_{x'}$ of $\mathbf{x'}$
		\IF{$\text{dist}(\mathbf{x'},\mathbf{nn}_{x'}) \leq \epsilon$}
		\IF{$\mathbf{nn}_{x'}$ belongs to \textit{noise} cluster}
		\STATE Create new cluster $C_{new}$ with samples $\mathbf{x'}$ and $\mathbf{nn}_{x'}$
		\STATE $C = C_{new}$
		\ELSE
		\STATE Assign $\mathbf{x'}$ to the same cluster $C_{nn}$ as $\mathbf{nn}_{x'}$
		\STATE $C = C_{nn}$
		\STATE PROPAGATE $C$ to $\epsilon$-neighborhood of $\mathbf{x'}$.
		\ENDIF
		
		\CGEMPTY
		\COMMENT{Process detected -- model adaptation}
		\IF{$|C| \geq minPts$}
		\STATE Train GMM $\mathcal{M}_{novel}$ of process $C$ with VI.
		\STATE Update $\mathcal{M}$ and fuse it with $\mathcal{M}_{novel}$.
		\STATE Remove $C$ and delete all samples of $C$ from $\mathcal{B}$
		\CGRETURN Classification of $\mathbf{x'}$ based on updated $\mathcal{M}$.
		\ENDIF
		\ENDIF		
		\CGRETURN classification of $\mathbf{x'}$ based on $\mathcal{M}$.
	\end{algorithmic}
\end{algorithm}
%

\subsubsection{Model Adaptation}

The last part of the 2SND procedure is responsible for deciding whether a \textit{novel} process exists in the monitored LDR and how the model is adapted.
If a new process in form of a cluster is identified, the underlying GMM needs to be updated. This is done by performing a VI training on all samples that are associated with the corresponding cluster. After that training step, the \textit{novel} process is represented by another GMM.
To update the model, we exploit the properties of the hyper-distributions and use a fusion technique proposed in \cite{Fis14}.
To fuse two given GMM we measure the pairwise divergence between each component and fuse only those which exceed a given threshold (0.5). In this case, we may assume that both components model the same process. This might happen, if a process emerges close to the border if the $\alpha$-region (which also separates LDR from HDR).
As divergence measure the Hellinger distance is used (cf. \cite{Bhat,Hel}). 
The actual fusion combines the hyper-parameters of both components.
Non-overlapping components are simply inserted into the existing model.
In each case the hyper-distributions of the mixing coefficients must be adjusted, such that
the mixing coefficients of the combined GMM still form a distribution.
After this fusion step, the model is adap\-ted to the changes in its environment and all samples belonging to the cluster $C$ are removed from buffer $\mathcal{B}$.
If the updated model is used as the base for a classifier, the conclusion for the new component must be determined. Possible solutions to this problem are the involvement of a human domain expert, if meaningful labels are required, or the automatic generation of new, unique labels.
As investigated by \cite{Fis12}, it is also possible to exchange knowledge with other systems so that a \textit{novel} process can be faster detected by another system. This kind of behavior is especially interesting for cyber-physical systems that share knowledge about their environment.

\begin{figure}[]
	\vskip 0.2in
	\begin{center}
		\subfigure[Initial training set with samples from two different classes, green circle $\circ$ and blue cross \textit{+}. The density model is trained with VI and extended to a classifier as described in Section \ref{sec:cmm}. \label{fig:ex:train}]{\includegraphics[width=.475\columnwidth]{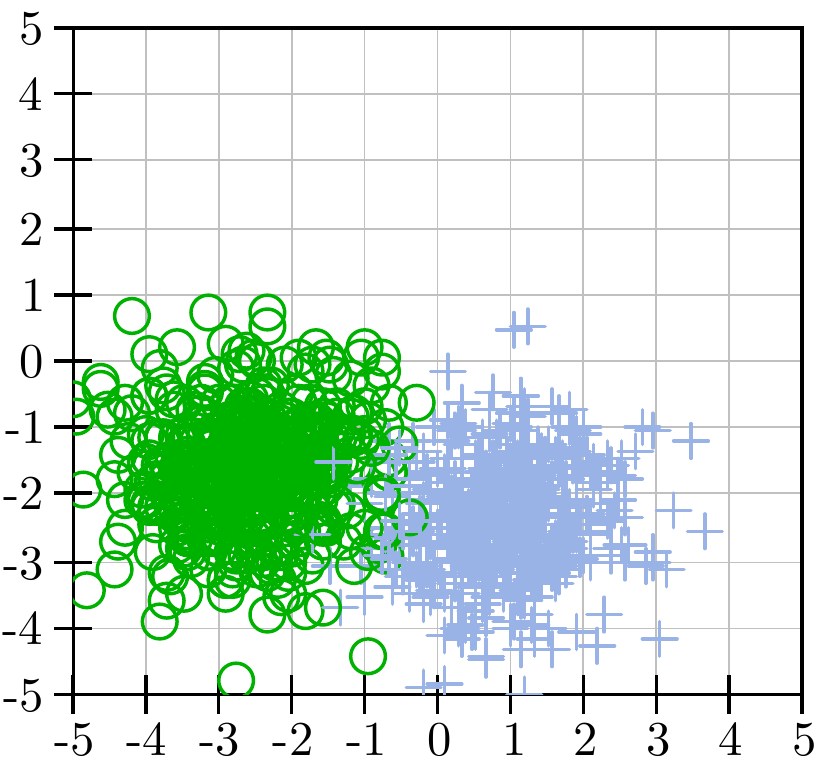}}
		\hspace{1mm}
		\subfigure[Resulting initial GMM with two components after VI training. The black line is the combination of the decision boundary and the $\alpha$-regions. Samples that appear in the outer (cyan colored) LDR are identified as \textit{suspicious}.\label{fig:ex:init}]{\includegraphics[width=.475\columnwidth]{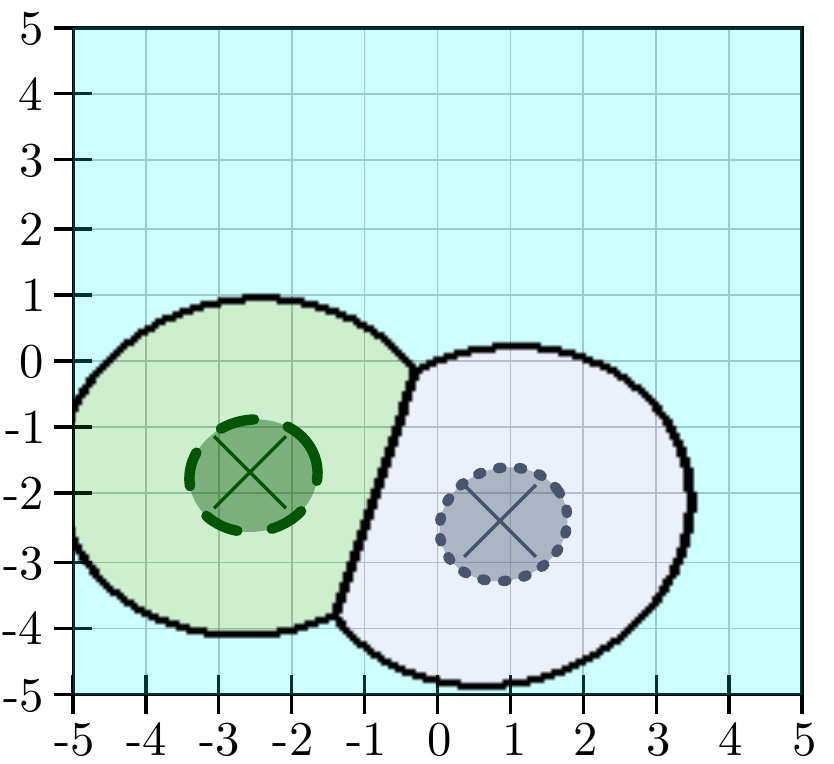}}
		\subfigure[Situation after the observation of \textit{potentially novel} samples. Different symbols represent samples of the same cluster, while blue triangles $\vartriangle$ are samples not yet assigned to a cluster.\label{fig:ex:detect}]{\includegraphics[width=.475\columnwidth]{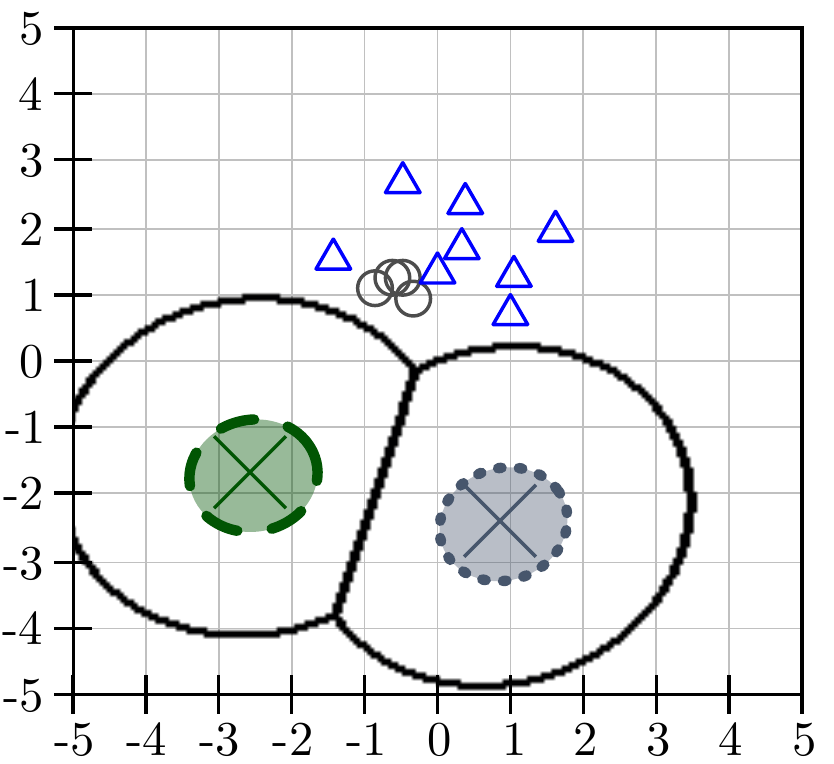}}
		\hspace{1mm}	
		\subfigure[After the apperance of some more \textit{suspicious} samples, the cluster in the upper center reached a certain size and is considered to be a \textit{novel} process. Its samples are isolated and used to train a parametric model with VI.\label{fig:ex:buffer}]{\includegraphics[width=.475\columnwidth]{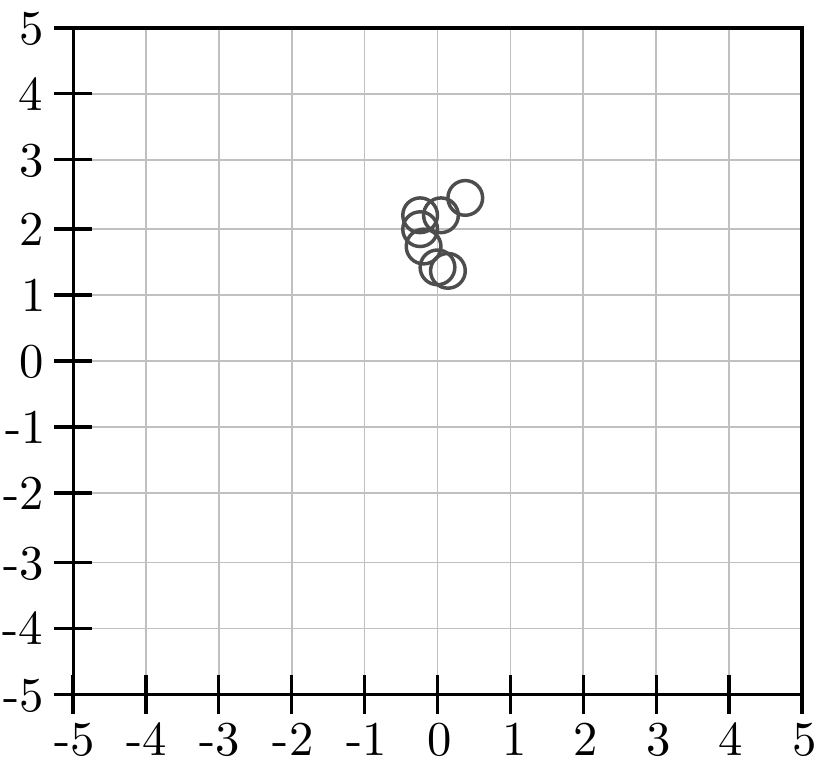}}	
		\subfigure[After the VI training converges, the \textit{novel} process is represented by a GMM which consists of a single component. The newly acquired knowledge will be fused with the initial model shown in \subref{fig:ex:init}. \label{fig:ex:proc}]{\includegraphics[width=.475\columnwidth]{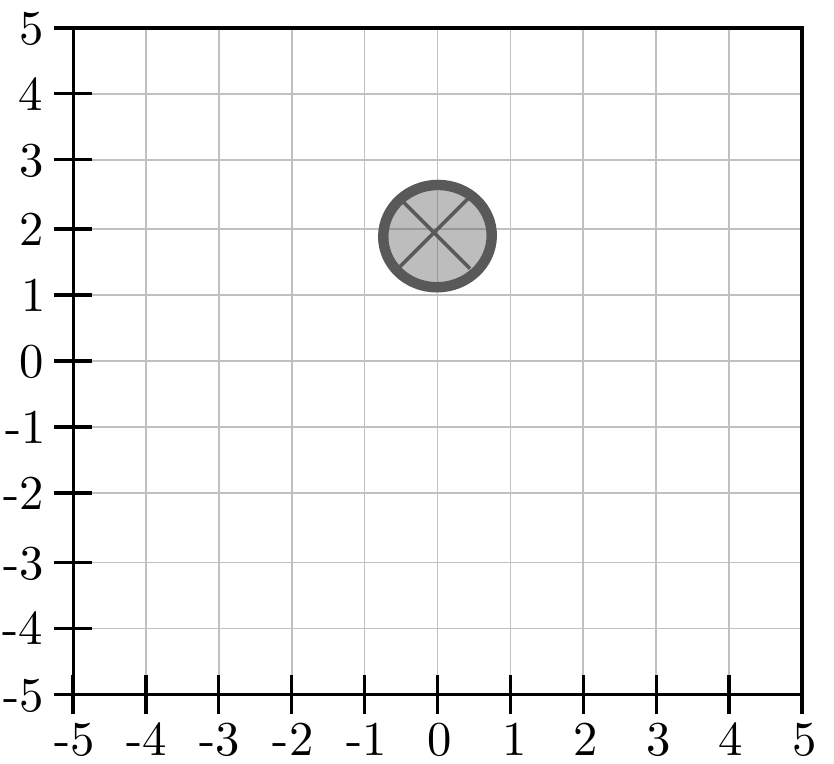}}	
		\hspace{1mm}
		\subfigure[Updated GMM and classifier after the \textit{novel} process is integrated. The updated decision regions are shown as well. The red component and region corresponds to the \textit{novel} process.\label{fig:ex:up}]{\includegraphics[width=.475\columnwidth]{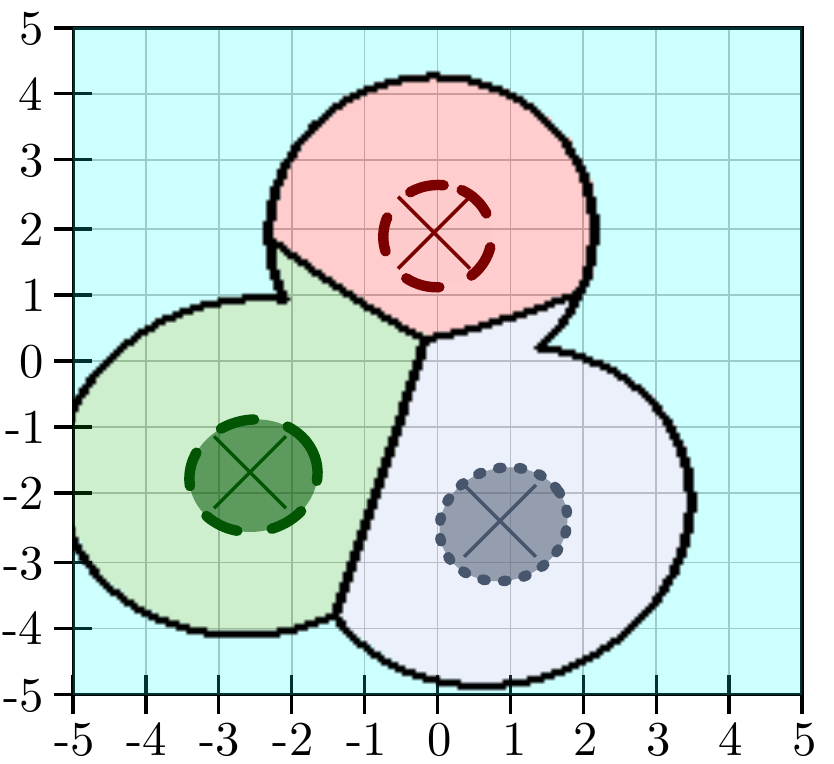}}	
		\caption{Illustration of the proposed technique.	
			In the training set only samples of two processes are present. In the operational phase a third process emerges and starts to generate samples. After enough \textit{potentially novel} samples are observed, the model and the classifier are updated.	\label{fig:examplary}}
	\end{center}
	\vskip -0.2in
\end{figure}

For clarification, an exemplary scenario that highlights the important operational phases of the new approach is shown in Figure \ref{fig:examplary}.

%
%

\subsection{Novelty Detection in High-Density Regions}
\label{sec:onlineGoF}

The above presented method detects \textit{suspicious} samples in LDR only. However, observations in HDR are  always considered as being \textit{normal}, thus making this approach unable to detect \textit{novel processes} there.
For the moment we focus on the detection of overlapping processes for single components and later extend the idea to also suit GMM (where multiple components are present).
The difficulty in detecting novel processes in dense regions \comment{oder HDR} is that we cannot decide if an observed sample is the legitimate outcome of a known (i.e.,~the existing component) or from an unknown overlapping process without knowing its affiliation (which in this case is a latent variable).
We therefore make use of a sliding window that keeps track of the last $\omega$ observed samples, no matter if they are actually \textit{novel} or not.
It is clear, that if a \textit{novel} process is present (that deviates at least in it's mean or covariance from the existing component) the observed sample population (i.e.,~the content of the sliding window) will not match the  distribution (described by the component) anymore and a noticeable difference between population and component has to be measurable.
Due to their high computational complexity divergence measures such as the Kullback-Leibler divergence \cite{KL51} or Hellinger distance \cite{Hel} are intractable for the measurements, especially if the input space is of high dimensionality.
To tackle this problem we do not measure the divergence of the sliding window and the existing component directly, instead we test how well the distances between samples in the buffer an the component's center suit the expected distribution (which is a $\chi^2$ distribution as stated in Section \ref{sec:2snd}). This task can be performed by using the $\chi^2$ test described in Section \ref{sec:found:chi2}. Since the test is performed on the sliding window, the approach is suitable for online environments.

\subsubsection{Transformation of the distance distribution}
\label{sec:transform}

One of the requirements of the $\chi^2$ goodness-of-fit test is, that the different events must be mutually exclusive. Therefore the continuous distance density must be transformed into a discrete one. Since any distribution is normalized (i.e.,~$\int p(x)\mathrm{d}x=1$) and, furthermore distances (here $x$) are always strictly positive it is quite easy to transform the density of distances into a discrete, uniform distribution.

Choosing a finite uniform distribution with $\lambda$ events (also cells, or buckets) brings several advantages including constant calculation of expected frequencies:
\begin{equation}
    \mathrm{unif}_{\lambda}(i) = \frac{1}{\lambda} ~ , ~ 1 \leq i \leq \lambda , i \in \mathbb{N}.
\end{equation}
The fitted uniform distribution has only one free parameter ($\lambda$), therefore the degree of freedom is given by:
\begin{equation}
    k = \lambda - 1.
\end{equation}
To do the actual transformation the boundaries of the individual cells ($\mathrm{cell}_i$) must be estimated so that each cell is equally likely. This is done with the inverse cumulative density function $F^{-1}$ of the continuous density (as stated before $\chi^2_d$ for multivariate Gaussians with $d$ dimensions) by dividing the density into $\lambda$ areas of equal size:
\begin{align}
    \mathrm{cell}_i &= 
     \begin{cases}
     [l_i,r_i)         & 1 \leq i < \lambda \\
     [l_i,\infty)         & i = \lambda
     \end{cases}, \\
    l_i &= F^{-1}_{\chi^2_d}\left((i-1) \cdot \frac{1}{\lambda} \right),\\
    r_i &= l_{i+1}.
\end{align}
Thus the first cell always begins at $0$, and the last one is unbounded (right boundary is $\rightarrow \infty$).

\begin{figure}
    \centering{
        \includegraphics[width=.49\columnwidth]{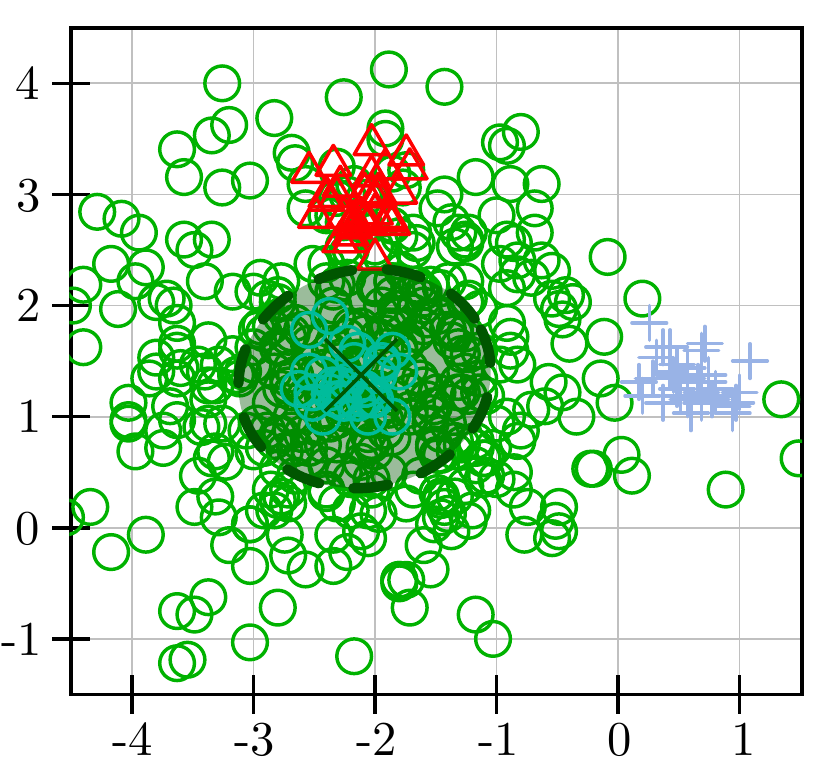}
        \includegraphics[width=.49\columnwidth]{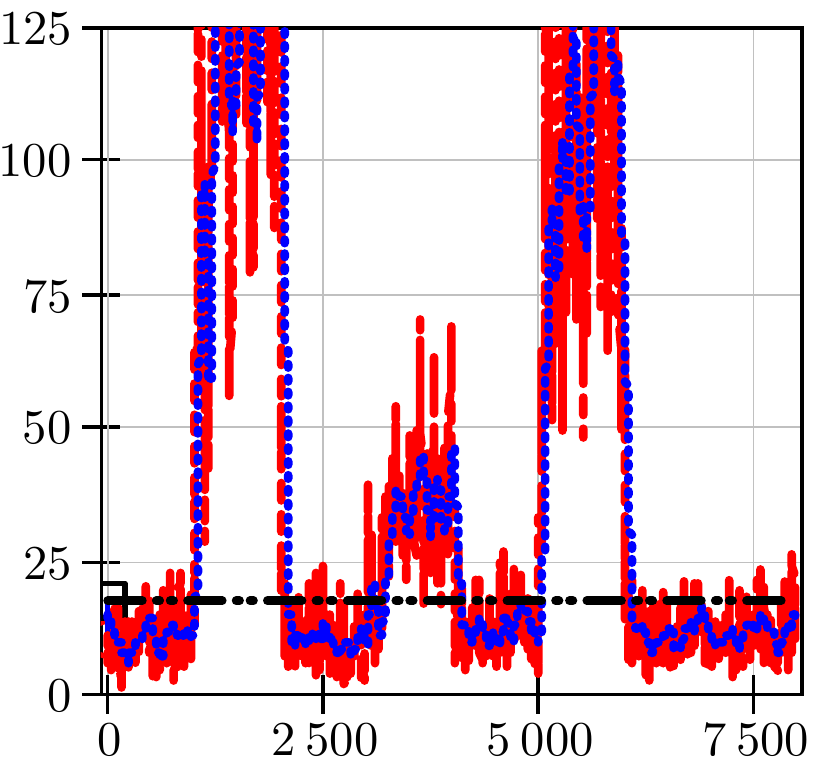}
    }
    \caption{Test data set and corresponding test statistics $t$. First crosses appear, then triangles, and finally circles. The parameters are: $w = 50$, $\lambda = 12$, $\alpha = 0.99$, $p = 0.01$. The red line is are the test statistics $t$. The horizontal, black line indicates the critical value.\label{fig:chi2detection}}
\end{figure}

Now, if a previously unseen sample $\bs{x'}$ is observed it is stored in the sliding window buffer and a counter $b_i$ for the responsible cell $\mathrm{cell}_i$ is incremented. The lookup for the correct cell can be done in $\mathcal{O}(\log n)$ (e.g.,~using a tree structure).
The $t$-value for the current buffer configuration at time point $n$ is then calculated as follows:
\begin{equation}
t_n = \sum_{i=1}^{\lambda} \frac{(b_{i,n} - e_i)^2}{e_i} \label{eq:tval}
\end{equation}
with
\begin{equation}
e_i = \mathrm{unif_\lambda}(i) \cdot \sum_{j=1}^{\lambda} b_j \label{eq:eval} \approx \frac{\omega}{\lambda}.
\end{equation}    
The expected value $e_i$ is only approximated by $\frac{\omega}{\lambda}$ since the buffer can be not completely filled (i.e.,~contains less than $\omega$ samples). If the buffer is at capacity when a new observation $\bs{x'}$ is processed, the oldest element gets removed. The $t$-value is compared with significance of $\alpha = 0.01 = 1\%$ against the critical value:
\begin{equation}
    \chi^{2,upper}_k = F^{-1}_{\chi^2_k}(1-\alpha) = F^{-1}_{\chi^2_k}(0.99).
\end{equation}
If $t > \chi^{2,upper}_k$ the threshold is exceeded and the detector reports \textit{novelty}.
Figure \ref{fig:chi2detection} shows on the left an exemplary data set with a single component trained to fit the green circles $\circ$. Additionally the set contains samples from three more processes (purple + crosses around $(0.5,1.5)$, red $\triangle$ triangles around $(-2,3)$, and blue $\circ$ circles close to the components center around $(-2,1.5)$). All additional processes represent \textit{novelties} and appear in the given order.
The image on the right shows the curve of the calculated $t$-values for the sliding window in red. When samples from the \textit{novel} processes appear the curve changes and exceeds the critical value (given as black line) considerably.

The signal is however noisy, so that at some points of time the critical value is slightly exceeded even though no \textit{novel} processes are present.
To compensate for this effect smoothing the $t$-values with a moving average:
\begin{equation}
t_{ma,n} = \frac{1}{M} \sum\limits_{i=0}^{M} t_{n-i},
\end{equation}
where $M$ is the number of considered previous $t$-values, is a promising approach as the blue curve on the right image of Figure \ref{fig:chi2detection} illustrates.	

\subsubsection{Learning of the distance distribution}
\label{sec:learned_dist}

\begin{figure}	
    \centering{
        \includegraphics[width=.49\columnwidth]{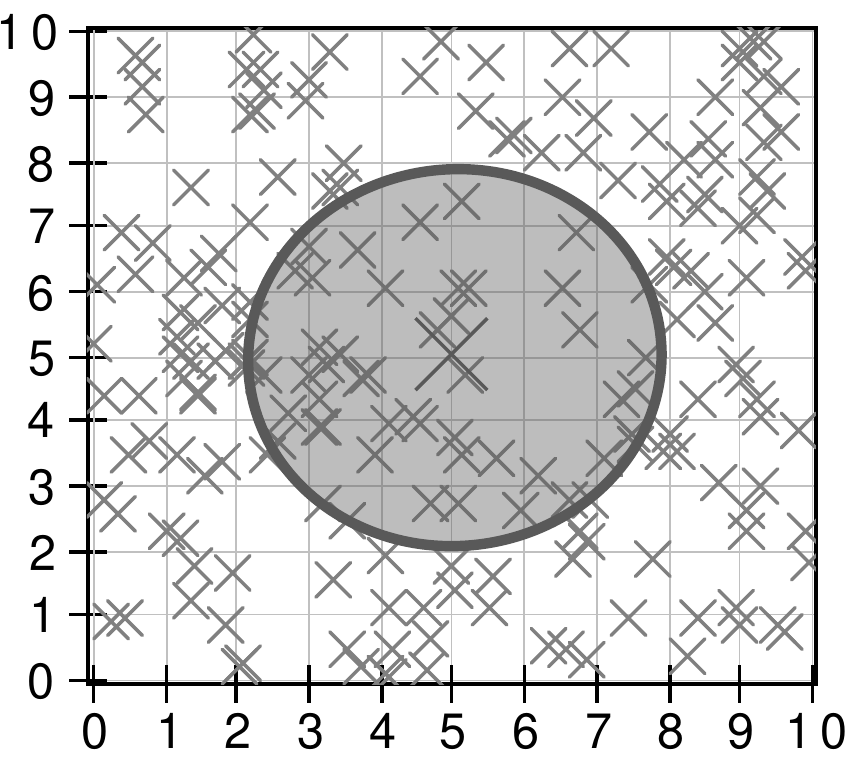}
        \includegraphics[width=.49\columnwidth]{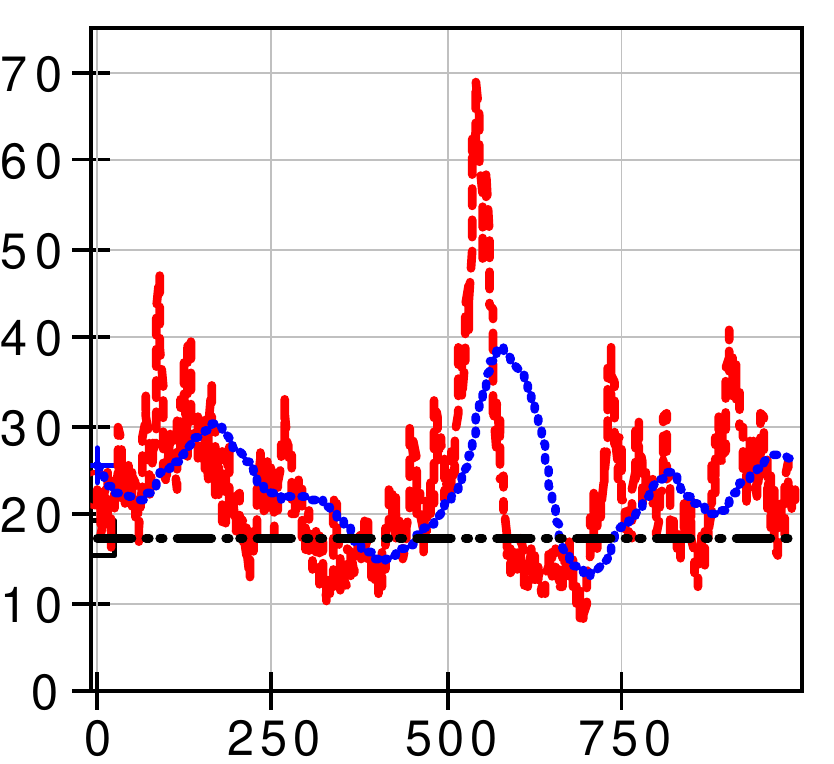}
    }
    \caption{1000 Uniformly distributed 2D samples in the interval $[0,10]$ and \textit{trained} Gaussian component. On the right the test statistics (red) and moving average test statistic (blue) of the whole applied train set. The critical value (black line) is clearly exceeded.\label{fig:distance_mismatch}}
\end{figure}

Real world data sets or sensory data from embedded systems often differ from their assumed distributions.
Whereas this is not a problem for classification for our goodness-of-fit approach to \textit{novelty detection} it is, as the $t$-value curve in Figure \ref{fig:distance_mismatch} highlights.
Here a single Gaussian is fitted based on the depicted samples, which are uniformly distributed rather than normal. Therefore the distances are not $\chi^2$ distributed as it is presumed by our test and the critical value is almost permanently exceeded by the test statistics.

The problem can be solved by estimating the cell boundaries directly from the samples $\mathcal{X}_{train}$ that are used to train the component:
\begin{align}
\mathrm{cell}_i &= 
\begin{cases}
[l_i,r_i)         & 1 < i < \lambda \\
[0,r_i)			  & i = 1 \\
[l_i,\infty)      & i = \lambda
\end{cases}, \label{eq:dist_estimate} \\
l_i &= \Delta( \bs{x}_{\lceil i \cdot\frac{w}{\lambda} \rceil}) \ , \ \bs{x}_j \in \mathrm{Sort}(\mathcal{X}_{train}),\\
r_i &= l_{i+1}.
\end{align}
where $\Delta(\bs{x})$ is the Mahalanobis distance to the center $\bs{\mu}$ of the component. By using the distance of every $\lceil i \cdot\frac{w}{\lambda} \rceil$ element of the order samples each cell contains approximately the same number of entries, thus forming again a discrete uniform distribution. The computationally most complex part is the sorting of the training samples $\mathcal{X}_{train}$, which can be done in $\mathcal{O}(n \cdot \log n)$. Note that the last $\mathrm{cell}_\lambda$ might be underestimated due to the rounding. 

\begin{figure}
    \centering
    \includegraphics[width=.66\columnwidth]{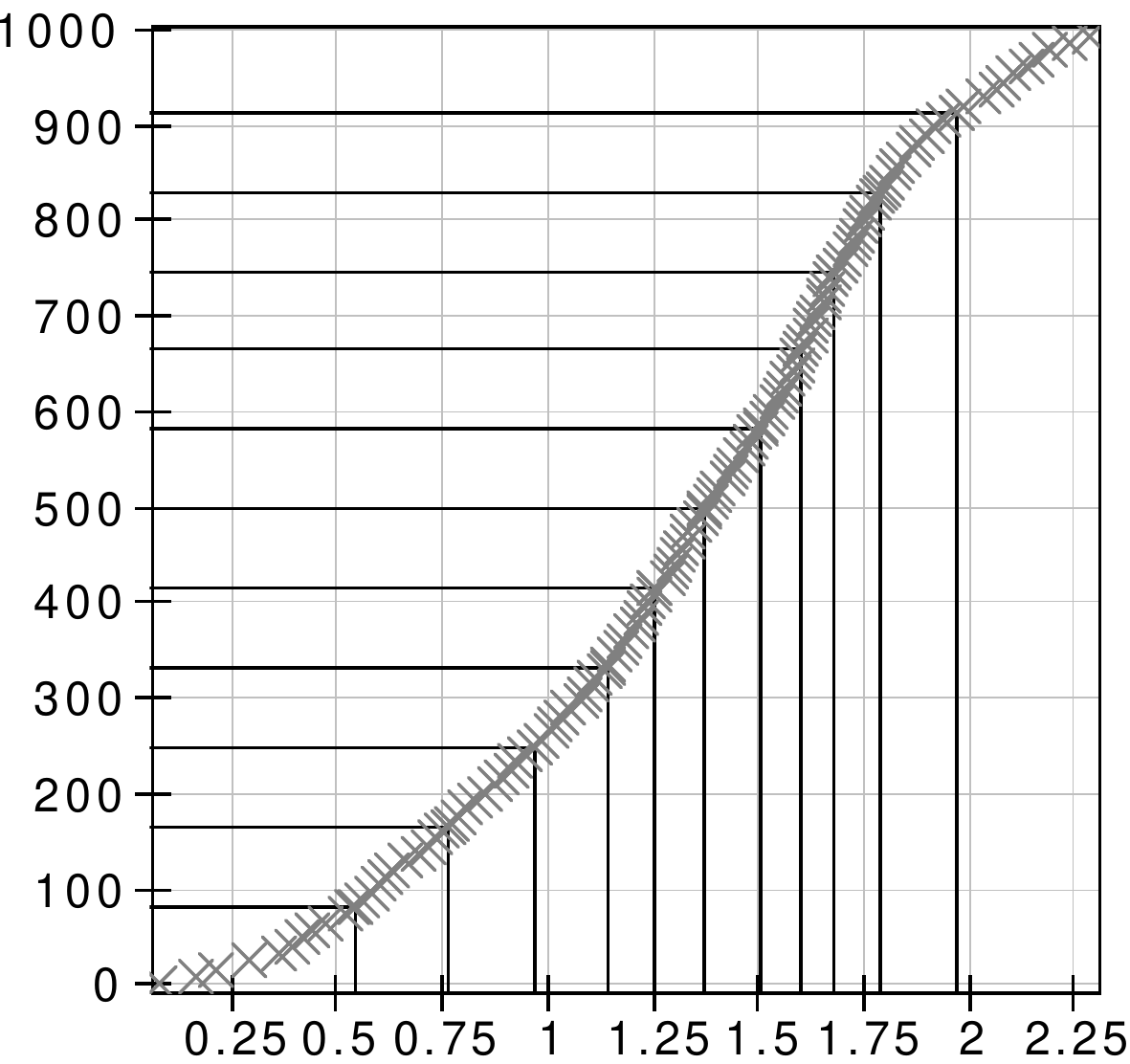}
    \caption{Ordered mahalanobis distances to the center of observed samples drawn from a bounded uniform distribution ($x \in [(x_1,x_2)| x_1,x_2 \in \mathrm{unif}(0,100)]$). The horizontal and vertical lines mark the boundaries for the cells of the resulting discrete uniform distribution.\label{fig:estunifdist} }
\end{figure}

In Figure \ref{fig:estunifdist} the ordered training set $\mathcal{X}_{train}$ is depicted. The $y$-axis (index of the sorted samples) is divided into $\lambda = 12$ equally sized parts (each part corresponding to a cell) the associated function arguments on the $x$-axis (distances to component center) are equal to the interval boundaries.
If the curve is normalized an approximation of the cumulative density function of the real distance distribution can be obtained.

\begin{figure}
    \centering{
        \includegraphics[width=.49\columnwidth]{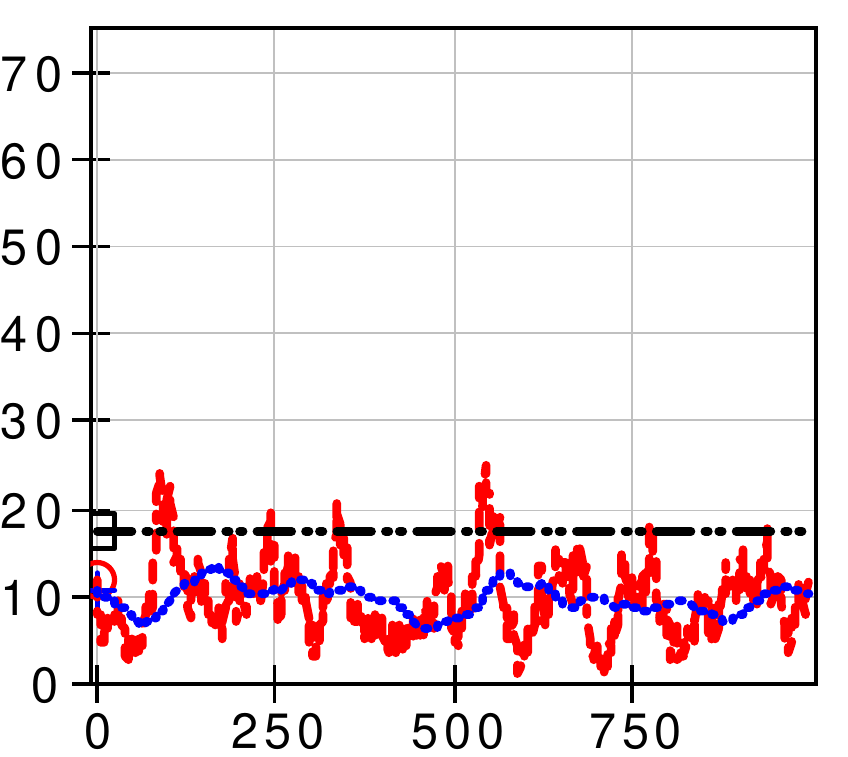}
        \includegraphics[width=.49\columnwidth]{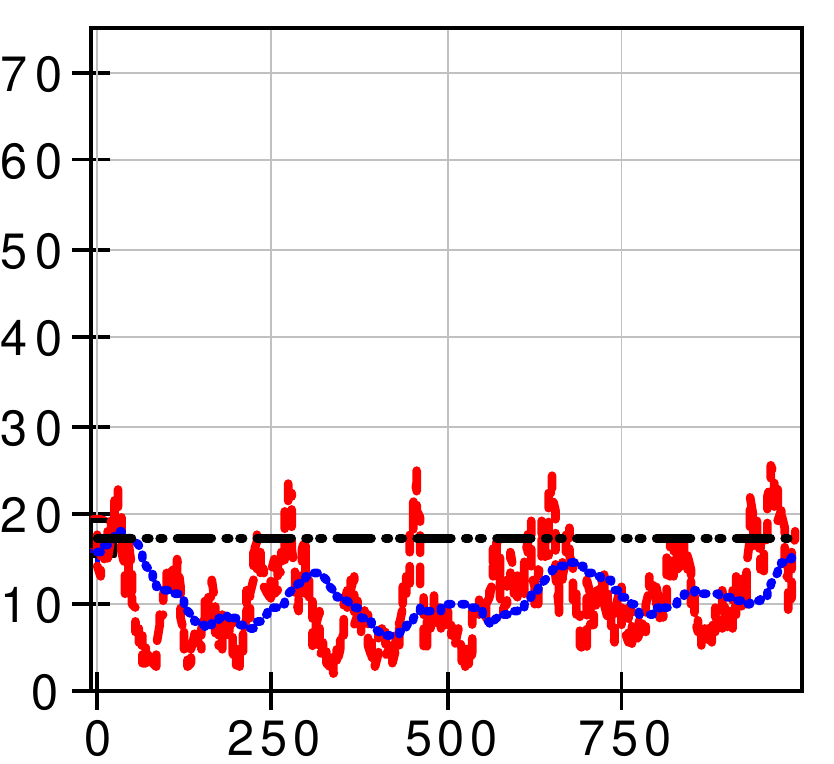}
    }
    \caption{Test statistics (red) and moving average test statistic (blue) each for uniformly distributed samples with estimated distance distribution. On the left side  for 1000 two dimensional samples, on the right for 1000 five dimensional samples. $w = 50$, $\lambda = 12$, $ma = 100$, $p = 0.01$\label{fig:distance_estimated}}
\end{figure}

Figure \ref{fig:distance_estimated} shows the resulting $t$-value curves of two uniform distributed data sets (on the left the same example as in Figure \ref{fig:distance_mismatch} with 2 dimensions, on the right another sample set with 5 dimensions) where the expected frequencies for the test are estimated according to Equation \eqref{eq:dist_estimate}.
The (moving-average) curves are now clearly below the critical value (black line), thus not indicating any \textit{novelty}.

\subsubsection{Extension to Gaussian Mixture Models}

\label{sec:onlineGoF:GMM}

To extend the high-density approach to GMM with multiple components, each component needs its own detector.

If a new sample $\bs{x'}$ is observed it should be used to update the detector of its affiliated component. The affiliation is however a latent variable and thus not known at run-time. One method to estimate the affiliations is (Monte Carlo) random sampling, which requires only the evaluation of the unnormalized (without mixing coefficient $\pi_j$) densities $P_j(\bs{x'}) = \mathcal{N}(\bs{x'}|\bs{\mu}_j,\bs{\Sigma}_j)$ for each component $j$ and a continuous uniform pseudo random number generator $\mathrm{unif}(0,1)$ for the unit interval $[0,1]$.
The sampling works by partitioning the unit interval into $J$ parts. Each partition $m_j$ is associated with exactly one component $j$ and the boundaries are given by:
\begin{align}
m_j &= 
\begin{cases}
[0,r_j)			  & j = 1 \\
[l_j,r_j)         & 1 < j < J \\
[l_j,1]      & j = J
\end{cases}, \label{eq:random_sampling} \\
p_{\bs{x'},j} &= \frac{P_j(\bs{x'})}{\sum_{k=1}^{J}P_k(\bs{x'})} \label{eq:random_sampling:norm}\\
l_j &= r_{j-1} &~ j > 1 ,\\
r_j &= l_j + p_{\bs{x'},j}.
\end{align}
where $p_{\bs{x'},j}$ are the normalized densities to ensure that the support of the individual parts sum up to $1$.
To identify a \textit{winner} component (i.e.,~the one that will be affiliated with the observation $\bs{x'}$) a random value $r'$ is drawn from the uniform generator $\mathrm{unif}(0,1)$. The partition $m_j$ that covers the drawn value $r'$ indicates the \textit{winning} component $j$.

\begin{figure}
    \subfigure[Clouds data set from UCI ML repo with trained CMM with 5fold cross validation.\label{fig:random_sampling:clouds}]{\includegraphics[width=.49\columnwidth]{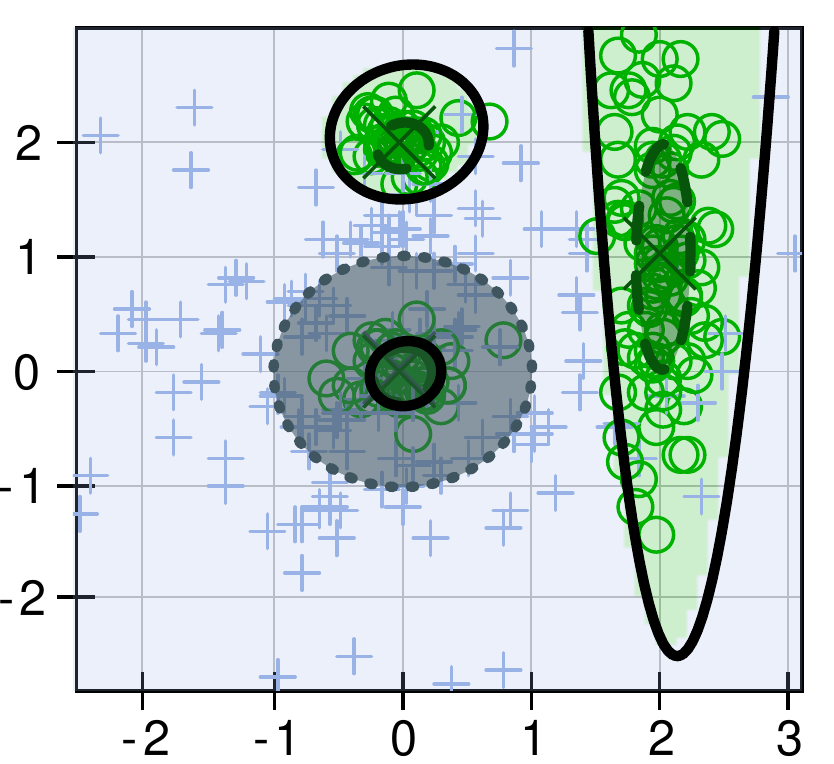}}		    \subfigure[Average novelty measures for high-density detectors. Clearly nothing suspicious is happening here.\label{fig:random_sampling:org_avg}]{\includegraphics[width=.49\columnwidth]{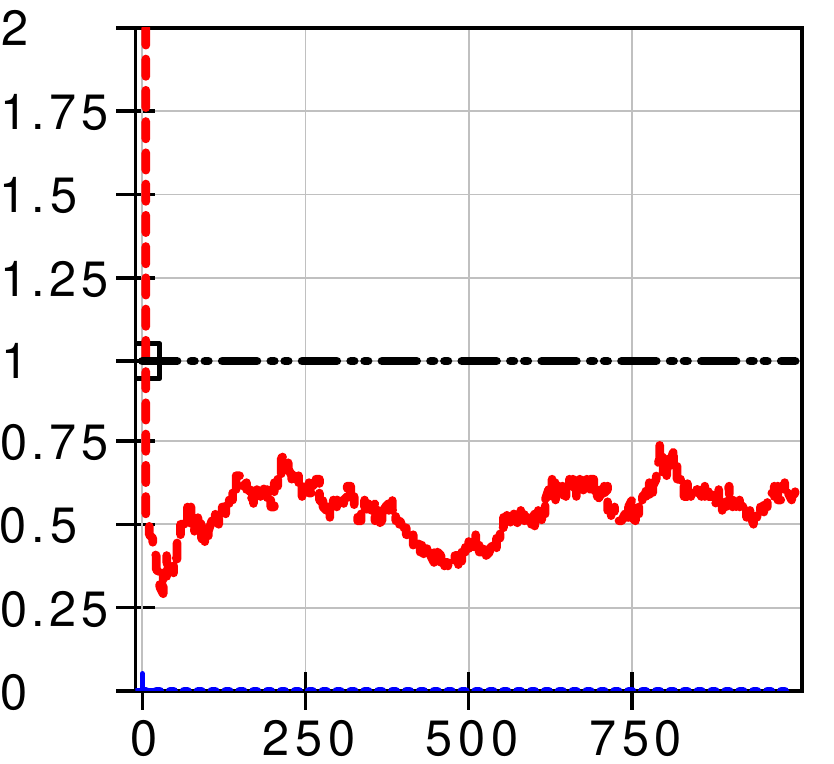}}	
    \caption{Data set with trained CMM and corresponding \textit{average-novelty} curve.}
\end{figure}

\begin{figure}
    \centering
    \subfigure{
        \includegraphics[width=.49\columnwidth]{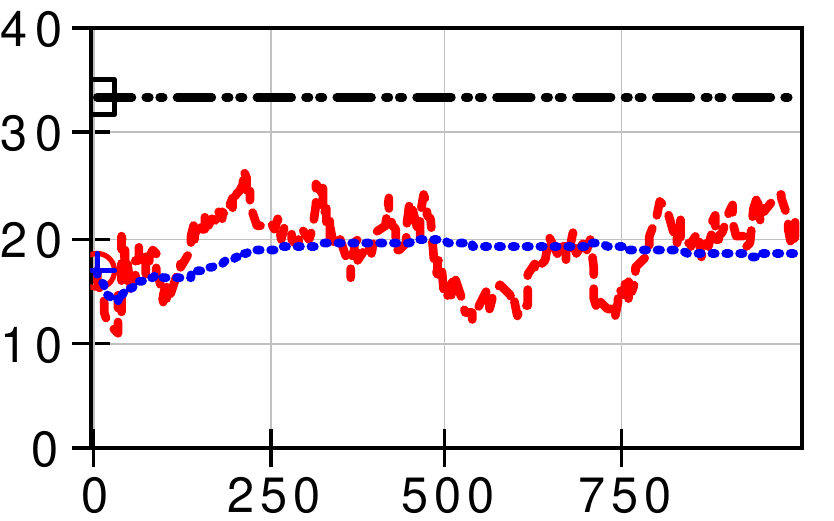}
        \includegraphics[width=.49\columnwidth]{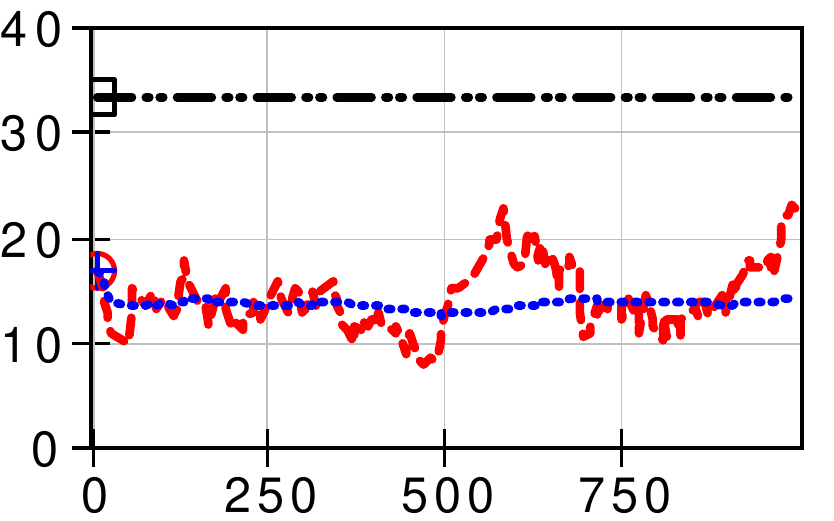}
    }                
    \subfigure{        
        \includegraphics[width=.49\columnwidth]{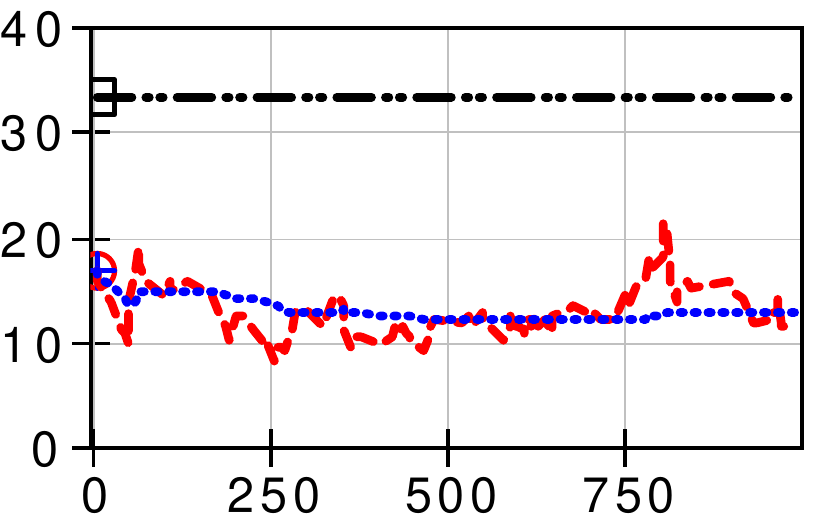}
        \includegraphics[width=.49\columnwidth]{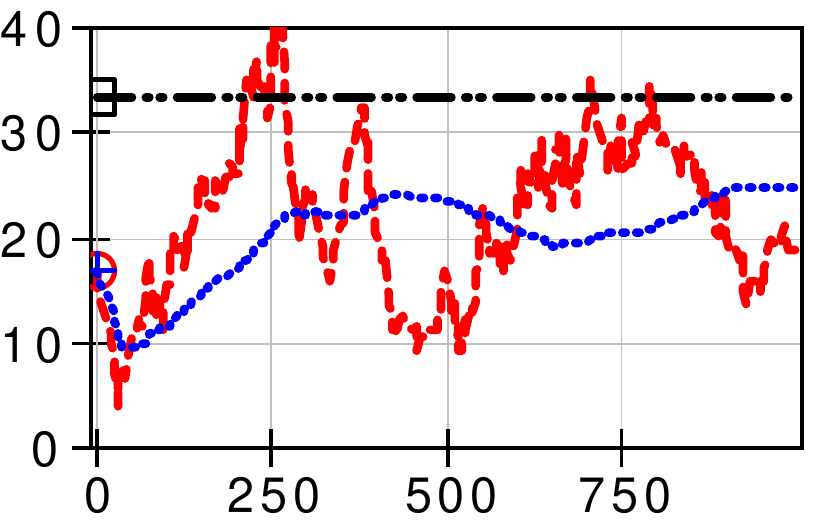}
    }
    \caption{Test statistics (red) and moving average test statistic (blue) each for remaining folds applied to all detectors. $w = 50$, $\lambda = 12$, $ma = 100$, $p = 0.01$\label{fig:random_sampling:org_detectors}}
\end{figure}

Figure \ref{fig:random_sampling:clouds} shows \textit{clouds} (5000 samples, 2 dimensions, 2 classes) a widely used artificial data set from the UCI Repository \cite{Lichman13}, with trained classifier.
The CMM is trained in 5-fold cross-validation fashion, where four folds are used for the actual training and the remaining fold for testing. This leads to an experiment where no \textit{novel} (unknown) processes are present since all folds contain samples from all four known processes.
The test result for one experiment is displayed in Figure \ref{fig:random_sampling:org_avg}. The curve shows the average high-density novelty measure (discussed in Section \ref{sec:measures}), which does not exceeded the critical value that is given by the black line and has a constant value of $1$, thus indicating that no \textit{novel} processes are present. The test statistics of the individual component detectors are depicted in Figure \ref{fig:random_sampling:org_detectors}.

\begin{figure}
    \subfigure[Samples drawn from clouds and two novel processes (red) which appear at ts $\approx 1000$ and $\approx 2000$.\label{fig:random_sampling:clouds_nov}]{\includegraphics[width=.49\columnwidth]{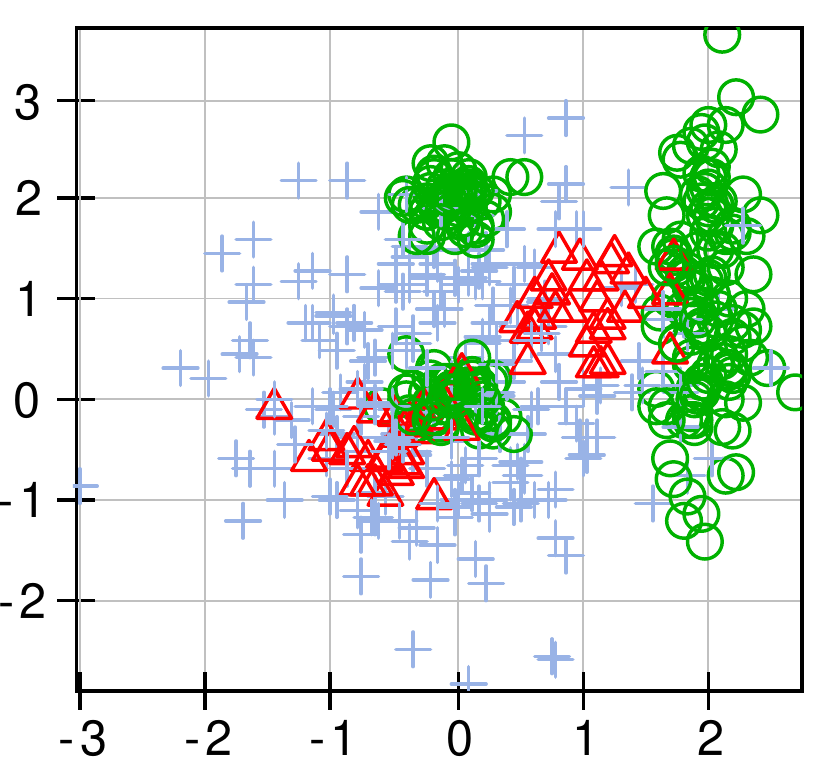}}		    \subfigure[Average novelty measure for high-density dectors. The blue line marks regions where a \textit{novel} process is detected.\label{fig:random_sampling:nov_avg}]{\includegraphics[width=.49\columnwidth]{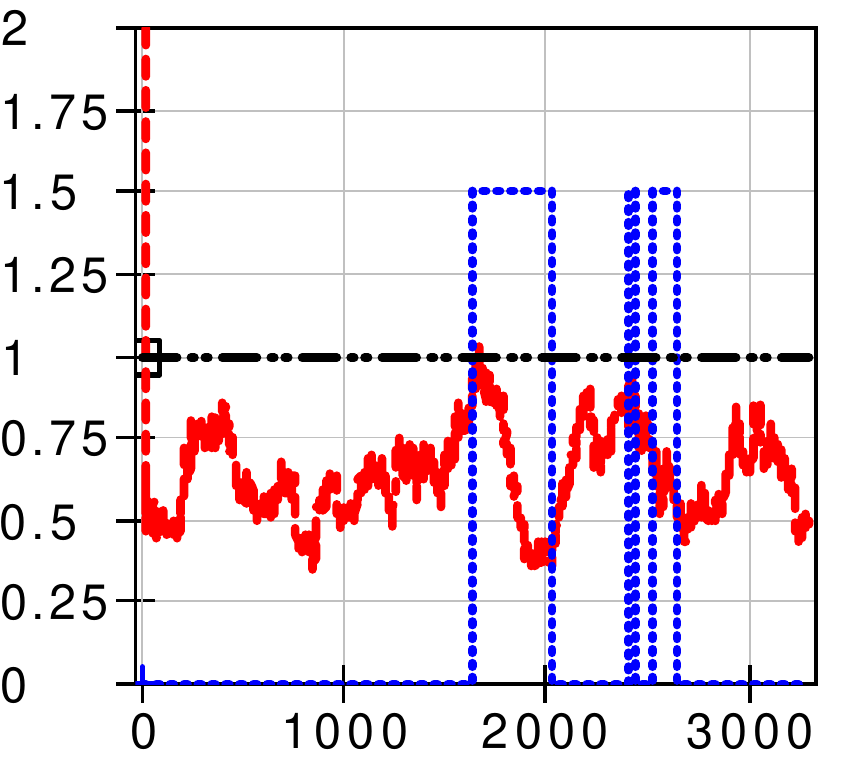}}	
    \caption{Data set with interspersed samples from \textit{novel} processes and corresponding \textit{average-novelty} curve.}
\end{figure}

\begin{figure}
    \centering
    \subfigure{
        \includegraphics[width=.49\columnwidth]{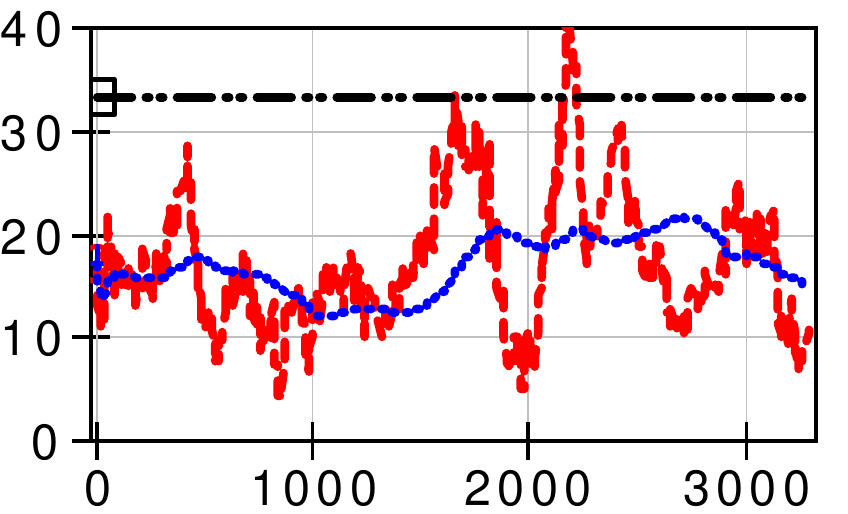}
        \includegraphics[width=.49\columnwidth]{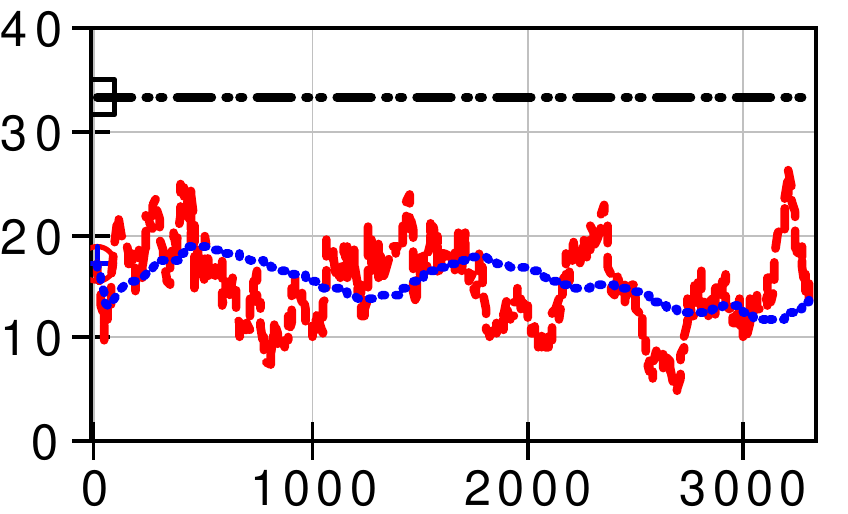}
    }                
    \subfigure{        
        \includegraphics[width=.49\columnwidth]{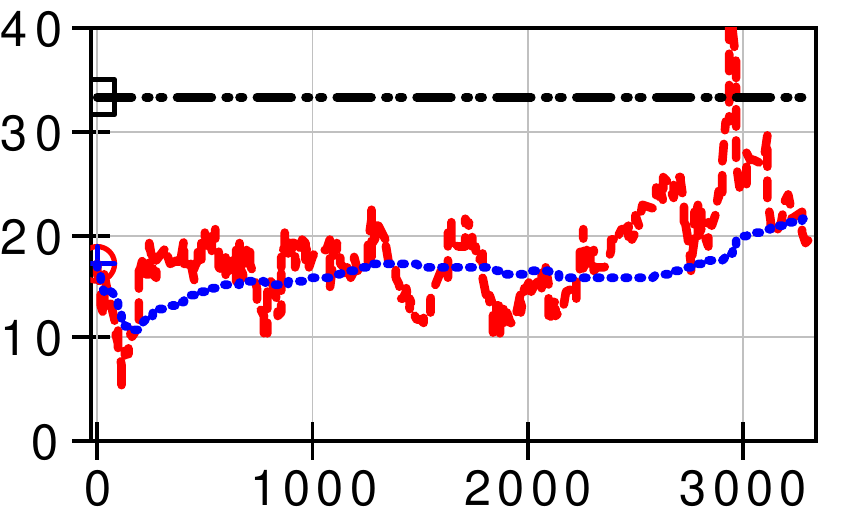}
        \includegraphics[width=.49\columnwidth]{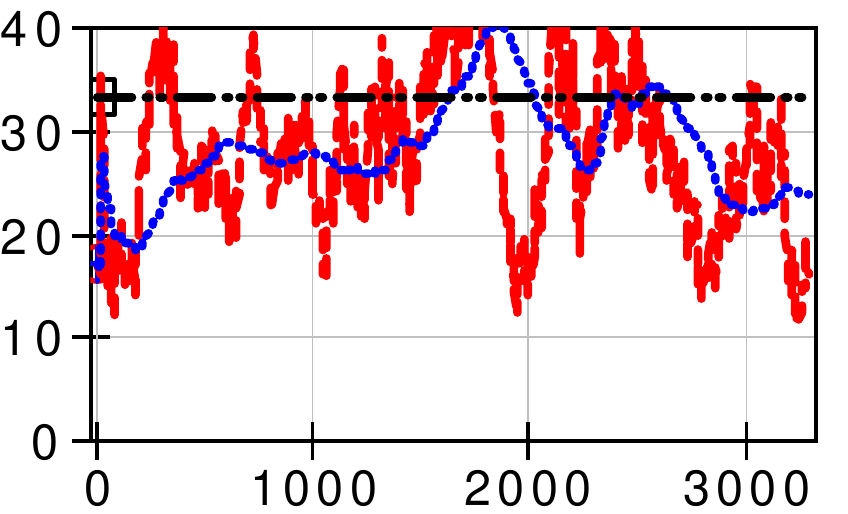}
    }
    \caption{Test statistics (red) and moving average test statistic (blue) each for modified clouds applied to all detectors. $w = 50$, $\lambda = 12$, $ma = 100$, $p = 0.01$\label{fig:random_sampling:nov_detectors}}
\end{figure}

A modified \textit{test} setup for clouds is illustrated in Figure \ref{fig:random_sampling:clouds_nov}. Here the training is performed on 2000 samples and the remaining 3000 samples are interspersed with 400 samples from two overlapping \textit{novel} processes (red $\triangle$ triangles). The novel processes appear around time steps 1000 and 2200. The corresponding high-density novelty measure curve for the experiment is given in Figure \ref{fig:random_sampling:nov_avg} and indicates \textit{novelty} (blue bars rising to $1$) in the regions where the \textit{novel} samples are interspersed.
The test statistics of the individual component detectors of this experiment are depicted in Figure \ref{fig:random_sampling:nov_detectors}. From the curves it can be inferred that the main contributions for the detection is from the component (bottom right) that represents blue + crosses.

\subsection{CANDIES}
\label{sec:candies}

CANDIES is our holistic approach to detect novelty in high- as well as in low-density regions of a GMM. This is achieved by using a single 2SND (cf. Section \ref{sec:2snd}) detector combined with multiple HDR detectors for each component (cf. Section \ref{sec:onlineGoF:GMM}).

\subsubsection{Requirements to merge LDR and HDR detectors}

For the system to operate some adjustments are necessary. At first a previously unseen sample $\bs{x'}$ is checked whether it is located in an HDR or not (similar to the first stage of 2SND). If this is not the case, the sample is passed to the second stage of 2SND and the density based clustering is refreshed. At this point a \textit{novel} process might be detected.
Otherwise, the in Section \ref{sec:onlineGoF:GMM} described random sampling is executed and $\bs{x'}$ gets affiliated with exactly one of the components $J$. Then for this component $j$ a new $t$-value is estimated and compared against the critical value. At this point an overlapping \textit{novel} process might be detected.

Since samples $\bs{x'}$ with a distance $\Delta_j(\bs{x'}) > \rho$ for all $j \in J$ are always processed by the LDR detection part, the assumed distance distribution of the components HDR detectors will not match the observed samples. However, by establishing the following dependency $\lambda = \frac{1}{1-\alpha}$ between the 2SND LDR detector and the HDR detectors, and adjusting the calculation of the $t$-values to:
\begin{align}
    t_{n,j} &= \sum_{i=1}^{\lambda-2} \frac{(b_{i,n} - e_i)^2}{e_i}, \label{eq:candies:tval} \\
e_i &\approx \frac{\omega}{\lambda-1} \label{eq:candies:eval},
\end{align}
the last cells $\mathrm{cell}_\lambda$ are representing exactly the fraction $\alpha$ of samples that are located in LDR (but with $e_\lambda = 0$) while the first $\lambda-1$ cells cover the remaining $1-\alpha$ percentage.
The critical value is changed to:
\begin{equation}
        \chi^{2,upper}_{\lambda-2} = F^{-1}_{\chi^2_{\lambda-2}}(0.99).
\end{equation}
Thus the goodness-of-fit test is adjusted to evaluate only the frequencies of samples expected to appear in HDR.
The whole approach is summarized and commented in Algorithm \ref{alg:CANDIES}.
\begin{algorithm}[tb]
    \caption{CANDIES}
    \label{alg:CANDIES}
   	\newcommand{\CGRETURN}{\STATE\textbf{return} }
    \newcommand{\CGEMPTY}{\STATE{\vspace{2.5mm}}}
    \begin{algorithmic}
		\STATE {\bfseries Input:} sample $\mathbf{x'}$, parameters $\alpha$, $\omega$ , $\epsilon$,$minPts$,$\tilde{b}$
        \STATE {\bfseries Global:} model $\mathcal{M}$, buffer $\mathcal{B}$
        \STATE Initialize $\rho =  F_{{\chi}_{D}^2}^{-1}(\alpha)$.
        
        \CGEMPTY
        \COMMENT{decide if $\bs{x'}$ is located in high- or low-density region}
        \FORALL{components $j$ in $\mathcal{M}$}
        \IF{$\Delta^2_j(\mathbf{x'}) \leq \rho$}
        \item \COMMENT{The observation is in dense region} 
        \item \COMMENT{get winner according to Section \ref{sec:onlineGoF:GMM}}
        \STATE{$j = $ winner component for $\bs{x'}$ }
        \STATE{update $\chi^2$ detector of component $j$}
        \item \COMMENT{Compare $t$-value with critical value} 
        \IF{$t_j > \chi^{2,upper}$}
        \item \COMMENT{Process detected} 
        \ENDIF
        
        \CGRETURN classification of $\mathbf{x'}$ based on $\mathcal{M}$.
        \ENDIF
        \ENDFOR       
        \item \COMMENT{The observation is in low-density region}               
        \CGRETURN 2SND$(\bs{x'},\alpha,\epsilon,minPts,\tilde{b})$.
    \end{algorithmic}
\end{algorithm}

\begin{figure}
	\centering
    \includegraphics[width=.49\columnwidth]{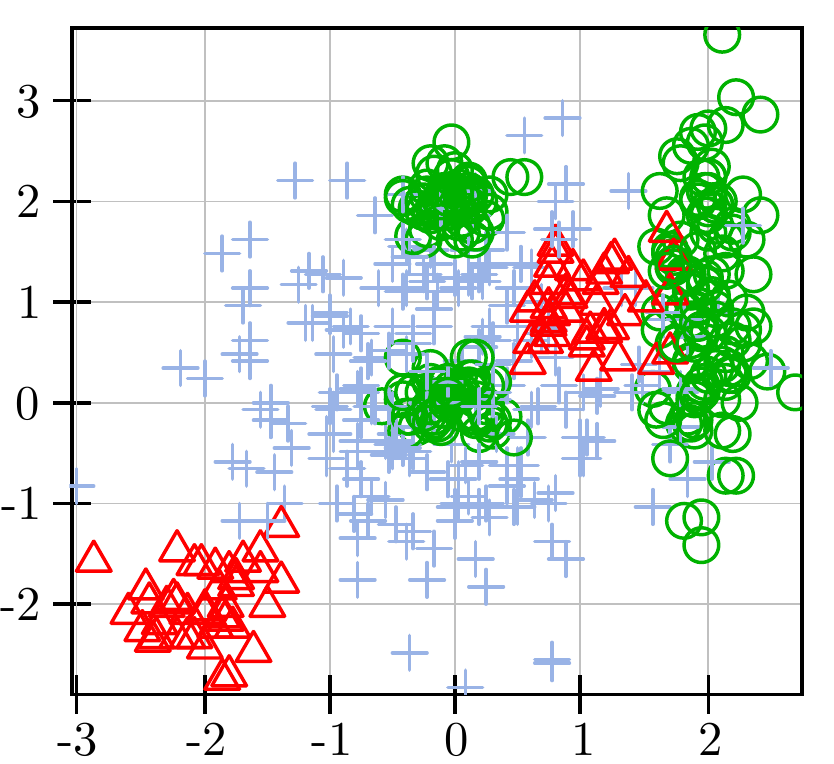}
    \caption{Samples drawn from clouds and two novel processes (red) which appear at ts $\approx 1000$ and $\approx 2000$.\label{fig:candies:clouds}}
\end{figure}

\begin{figure}
	\centering
    \subfigure{
        \includegraphics[width=.49\columnwidth]{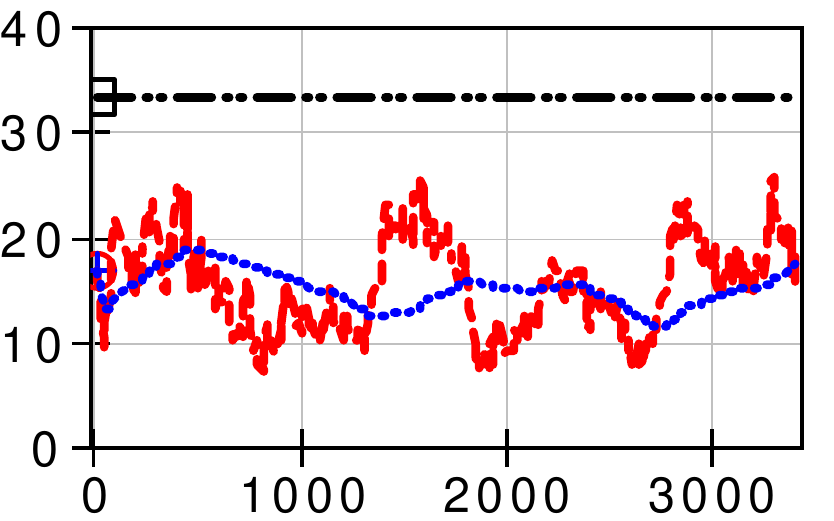}
        \includegraphics[width=.49\columnwidth]{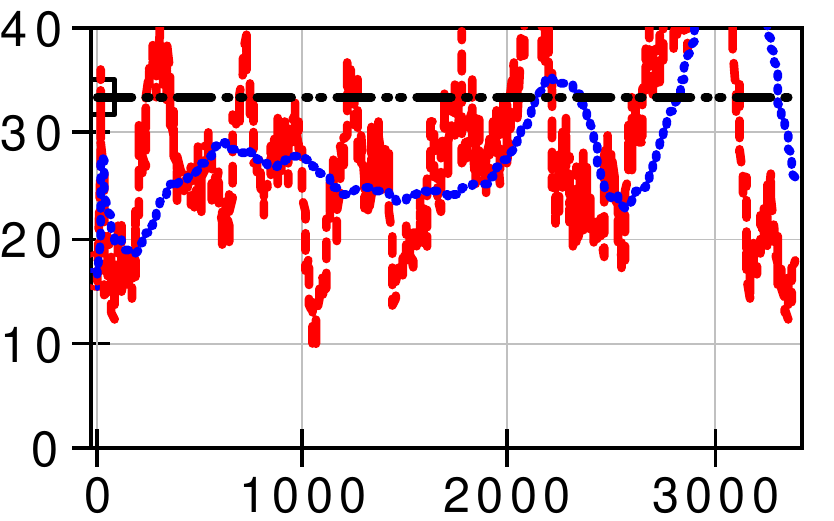}
	}                
    \subfigure{        
        \includegraphics[width=.49\columnwidth]{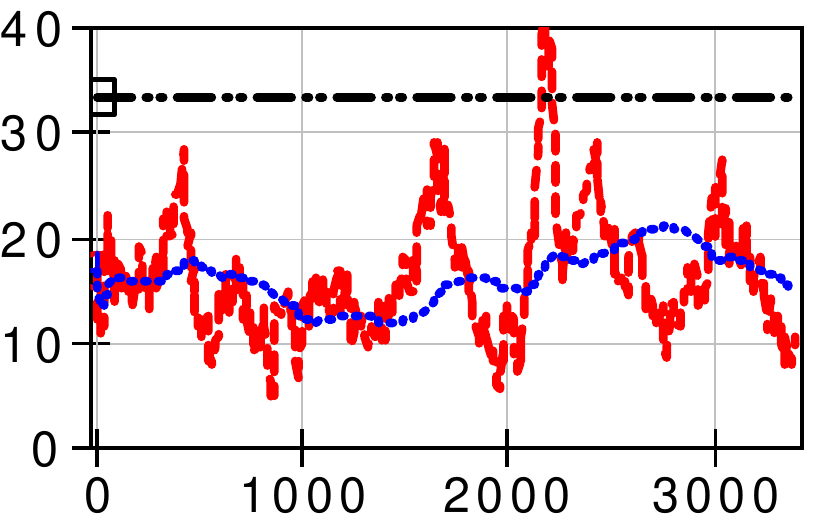}
        \includegraphics[width=.49\columnwidth]{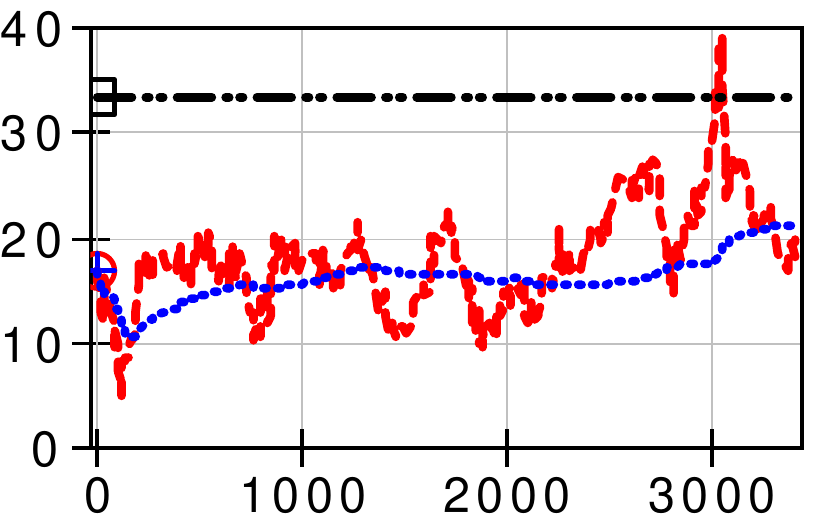}
    }
    \caption{Test statistics (red) and moving average test statistic (blue) each for modified clouds applied to all detectors.\label{fig:candies:gmm}}
\end{figure}

\begin{figure}
	\centering
    \subfigure{
        \includegraphics[width=.49\columnwidth]{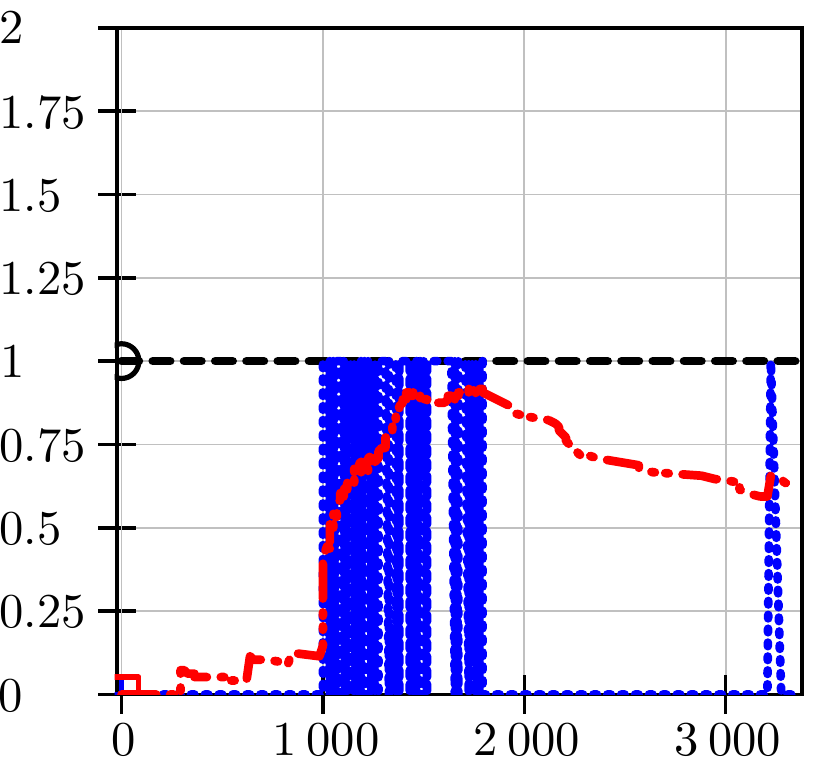}
        \includegraphics[width=.49\columnwidth]{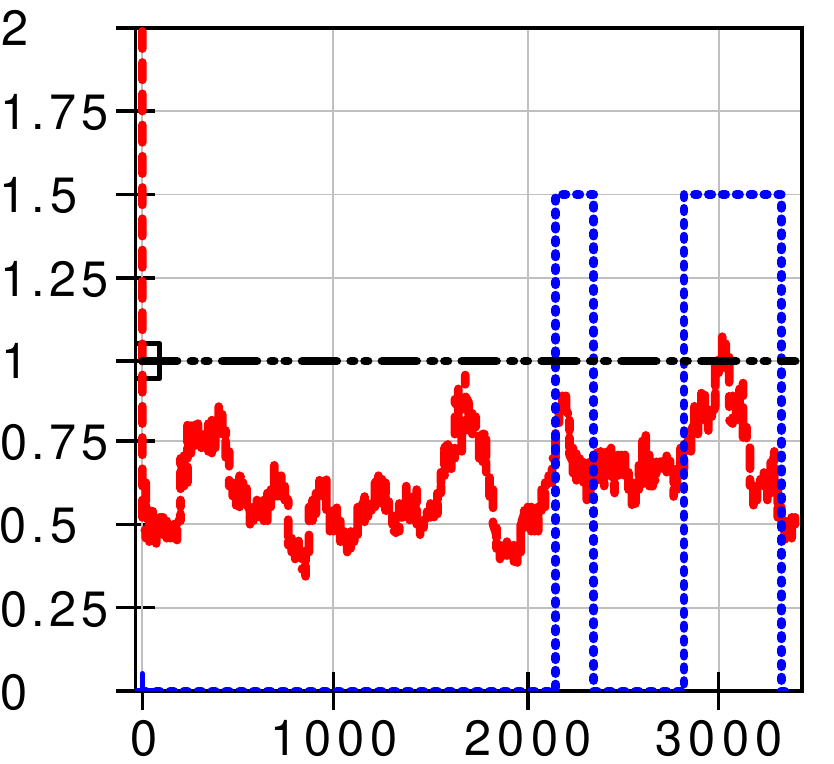}
	}                
    \caption{Measured novelty for modified clouds (cf. Figure \ref{fig:candies:clouds}).On the left: \textit{2SND-novelty} curve indicating a \textit{novel} process in low-density regions around ts $\approx1000$. On the right: \textit{average-novelty} curve that represents \textit{novelty} in high-density regions. The rising of the blue bars around ts $\approx2150$ and $\approx2900$ are indicating that a \textit{novel} process is present.\label{fig:candies:measures}}
\end{figure}

Figure \ref{fig:candies:clouds} shows another modification of the previously used \textit{clouds} data set. Again, additional samples (red $\triangle$ triangles) from two \textit{novel} processes are interspersed. The locations are chosen in a way so that one process (centered around $(1,1)$) shares a large fraction of its support with two of the known processes, while the other one (centered around $(-2,-2)$) is positioned in a low-density region.
The $t$-value curves of the four \textit{known} components are illustrated in Figure \ref{fig:candies:gmm}.
 Furthermore the \textit{novelty measures} (discussed in Section \ref{sec:measures}) given in Figure \ref{fig:candies:measures} clearly indicate \textit{novelty} in the expected time ranges (blue bars rising to $1$). The left curve gives the novelty value in low-density regions, where the first \textit{novel} process starts generating samples around time stamp $\approx1000$. On the right curve the \textit{average novelty} measure for high-density regions is illustrated. Here, the \textit{novel} process gets also detected but a small delay between appearance of \textit{novel} samples (around ts $\approx 2000$) and the detection can be observed. This is most likely due to the \textit{random sampling}, which disperses \textit{novel} samples to multiple components.

\subsubsection{Novelty Measure -- Human Readability}
\label{sec:measures}

We propose two \textit{novelty measures} to quantify how much \textit{novelty} is present in different regions (i.e. HDR or LDR) in a way that is comprehensible for (data scientists). Therefore, the measures should express the absence of a \textit{novel} process (or \textit{novelty}) with a value near $0$, while the presence of such a process should be expressed by a value $\geq1$.

The measure $\nu_{2snd}$ for LDR is given by:
\begin{equation}
    \nu_{2snd,n} = 1 - \frac{|\mathcal{C}|+|Noise|}{|\mathcal{B}|}
\end{equation}
Where $|\mathcal{B}|$ is the number of observations currently stored in the buffer, $|\mathcal{C}|$ is the number of different cluster, and $|Noise|$ the number of samples associated with the \textit{noise} cluster.
If a single cluster that contains most samples currently kept in the buffer is present (which is a strong indicator for a novel process), the measure will be close to $1$.
On the other hand, if all samples are considered to be \textit{noise} or multiple clusters with only a few samples are present, the \textit{novelty} value will be closer to $0$.

The HDR measure $\bar{\nu}$ (\textit{average novelty}) is based on the geometric mean of the normalized $t$-values $\nu_j$ of the individual components:

\begin{align}
    \bar{\nu}_n &= \left( \prod_{j}^{J} w(\nu_j) \right) ^ \frac{1}{J}, \label{eq:avg-nov} \\
    \nu_j &= \frac{t_{n,j}}{\chi^{2,upper}_{k}}.
\end{align}
The normalization constant is given by the critical value. As Equation \eqref{eq:avg-nov} shows, the $\nu_j$ are passed to a function $w$ which is a non-linear transform that boosts values near $1$:
\begin{align}
    w(x) &= x \cdot (2-\mathrm{comp}(1-x,1000)), \label{eq:weight_compressor} \\
        \mathrm{comp}(x,\mu) &= \frac{\log(1+\mu\cdot x)}{\log(1+\mu)}.
\end{align}
The idea here is that if multiple components approach the critical value (an indicator for \textit{novel} process located between these components) the \textit{novelty measure} should also express this. If the model however consists of considerably more components (with $t$-values distant from the critical value), the mean is dominated by these components. Thus boosting values already close to $1$, allows to overcome the \textit{normal} components to increase the mean, so that \textit{novelty} is also expressed there.
\begin{figure}
	\centering
    \subfigure{
        \includegraphics[width=.49\columnwidth]{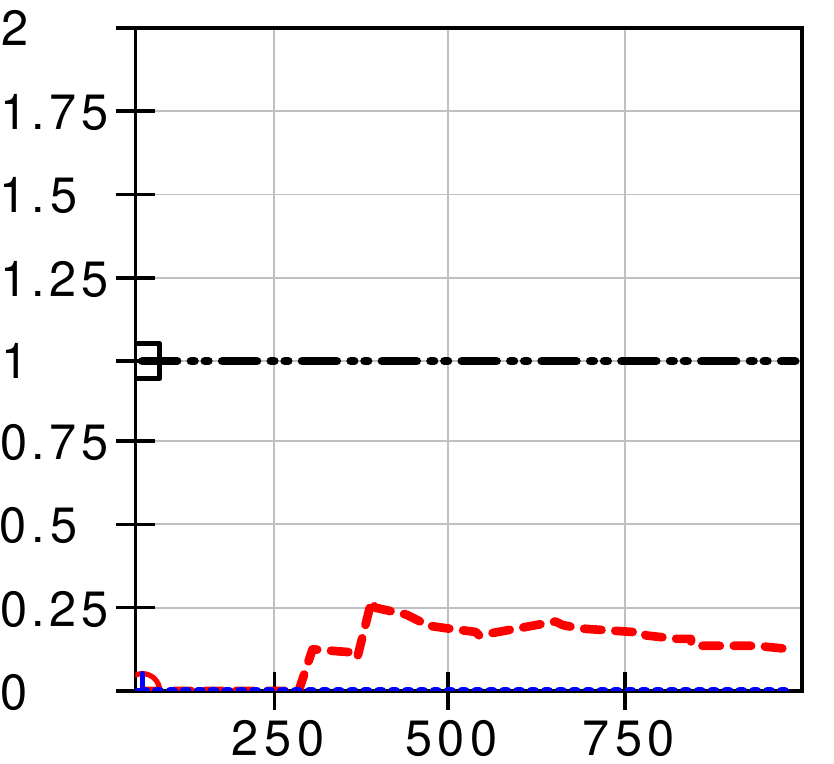}
        \includegraphics[width=.49\columnwidth]{clouds_org_avg}
	}                
    \caption{Measured novelty for clouds.On the left: \textit{2SND-novelty} curve showing low \textit{novelty} values for low-density regions. On the right: \textit{average-novelty} curve that represents \textit{novelty} in high-density regions.\label{fig:measures}}
\end{figure}
Exemplary curves for both measures are given in Figure \ref{fig:candies:measures} (\textit{novel} processes present) and Figure \ref{fig:measures} (only \textit{normal} processes observed).

\subsubsection{Overview of Parameters}

\theApproach~ comes along with a considerable amount of adjustable parameters. Table \ref{tab:params} gives an overview of all parameters present in \theApproach~ including a short description, recommendations for (good) default values (if possible), and which detector is influenced by the parameter.
\begin{table*}[]
    \centering
    \caption{Different parameters necessary for the proposed combined novelty detection approach.}
    \label{tab:params}
    \begin{tabular}{l|lll}
        Parameter               & Description                                                                                                  & Default              & Detector     \\
        \hline         
        \hline
        $\omega$                & \begin{tabular}[c]{@{}l@{}}Window size per component detector\end{tabular}                           &         $5\cdot\lambda$             &  High-Density        \\
        $\lambda$               & Number of discrete levels                                                                                     & $\frac{1}{1-\alpha}$ &  High-Density        \\
        ma                      & Size of MovingAverage filter                                                                                 & $2 \cdot \omega$     &  High-Density        \\
        p                       & $\alpha$-Value for $\chi^2$ test                                                                                    & 0.01                 &  High-Density        \\
        $\alpha$                & Size of alpha region                                                                                         & 0.95                 & Low-Density: 1. Stage \\
        $\epsilon$              & \begin{tabular}[c]{@{}l@{}}Maximum distance between\\ samples in a cluster\end{tabular}                      & 2                    & Low-Density: 2. Stage \\
        $|\mathcal{B}|$         & \begin{tabular}[c]{@{}l@{}}Buffer for Samples in low \\ density region\end{tabular}                          & 100                  & Low-Density: 2. Stage \\
        $P(\mathcal{C})$=minPts & \begin{tabular}[c]{@{}l@{}}Size of a cluster to be considered\\ as the outcome of a new process\end{tabular} & 10                   & Low-Density: 2. Stage
    \end{tabular}
\end{table*}
Note, that especially the buffer-size parameters are application-dependent on how many \textit{novel} processes are expected to appear at once, and how many samples they will generate.

\subsubsection{Handling of Noise}

Since the novelty detection is designed to detect \textit{novel} processes and not single observations, it is rather robust against distributed noise in the input space. 
While \textit{novel} samples of a \textit{novel} process will appear in a dense form, random noise is scattered across the input space so that it is quite unlikely to form sufficiently large clusters.

\begin{figure}[h]
	\vskip 0.2in
	\begin{center}
		\subfigure[Test set consisting of samples from one previously known process (green circles $\circ$), noise samples (red triangles $\vartriangle
		$) and samples from a novel process (blue crosses \textit{+}) .\label{fig:noise:data}]{\includegraphics[width=.475\columnwidth]{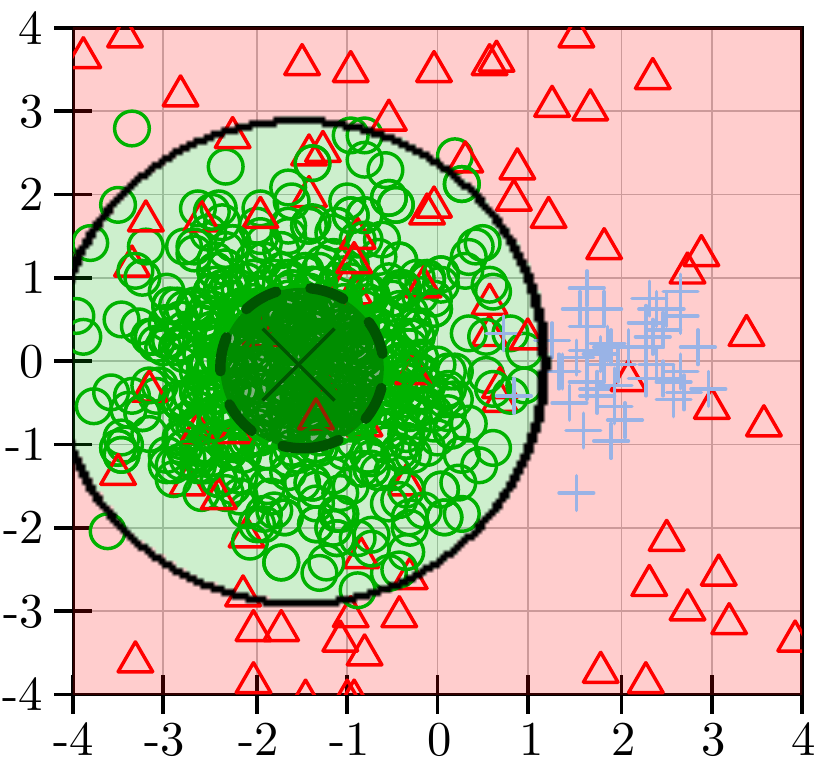}}	
		\hspace{1mm}
		\subfigure[Model identified by our approach after the process (blue $\times$ surrounded by dotted ellipse) that was responsible for the blue crosses is detected and integrated into the classification model.\label{fig:noise:cmm}]{\includegraphics[width=.475\columnwidth]{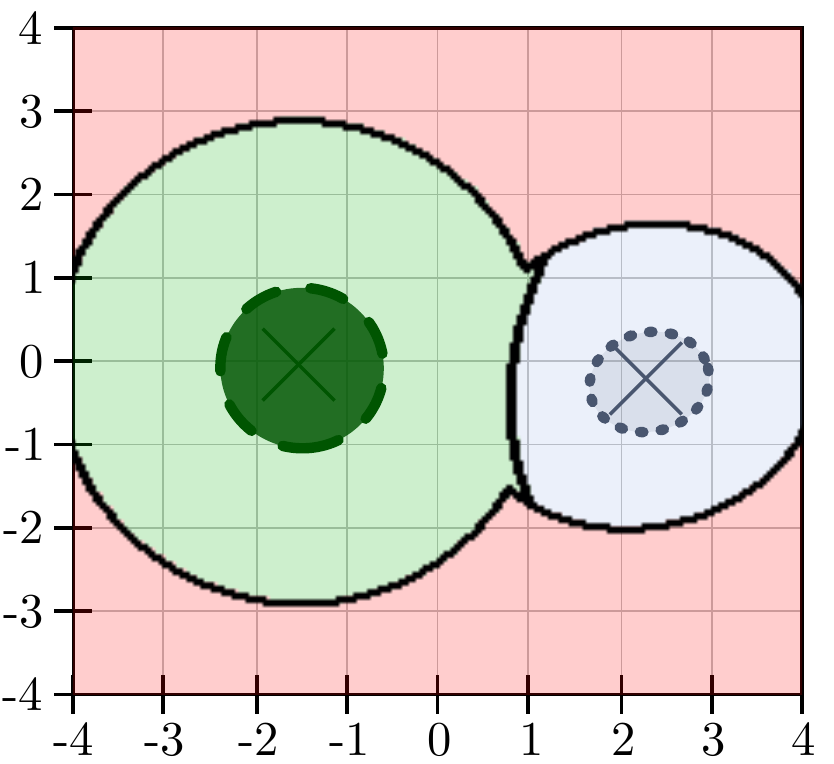}}	
		\caption{Scenario with samples from a \textit{novel} process and additional uniform distributed noise that is scattered in the input space. The region where observations are identified as \textit{novel} is colored in red, regions with different class assignments are also separated by a solid black decision boundary.\label{fig:noise}}
	\end{center}
	\vskip -0.2in
\end{figure}

For the LDR detection part (based on 2SND) the robustness is achieved by the two stage architecture that \textit{suspicious} samples pass through. 
Figure \ref{fig:noise} shows a scenario that includes uniformly distributed noise that is mixed into a test set with observations from one known process and one novel process (located to the right, outside of the $\alpha$-zone in low-density region). Depending on its parametrization, the LDR approach only detects a \textit{novel} process where the \textit{novel} observations are actually located.

\begin{figure}[h]
    \vskip 0.2in
    \begin{center}
        \subfigure[The same data set as in Figure \ref{fig:chi2detection} but with uniform noise added (purple crosses \textit{+}).\label{fig:noise_candies:ds}]{\includegraphics[width=.66\columnwidth]{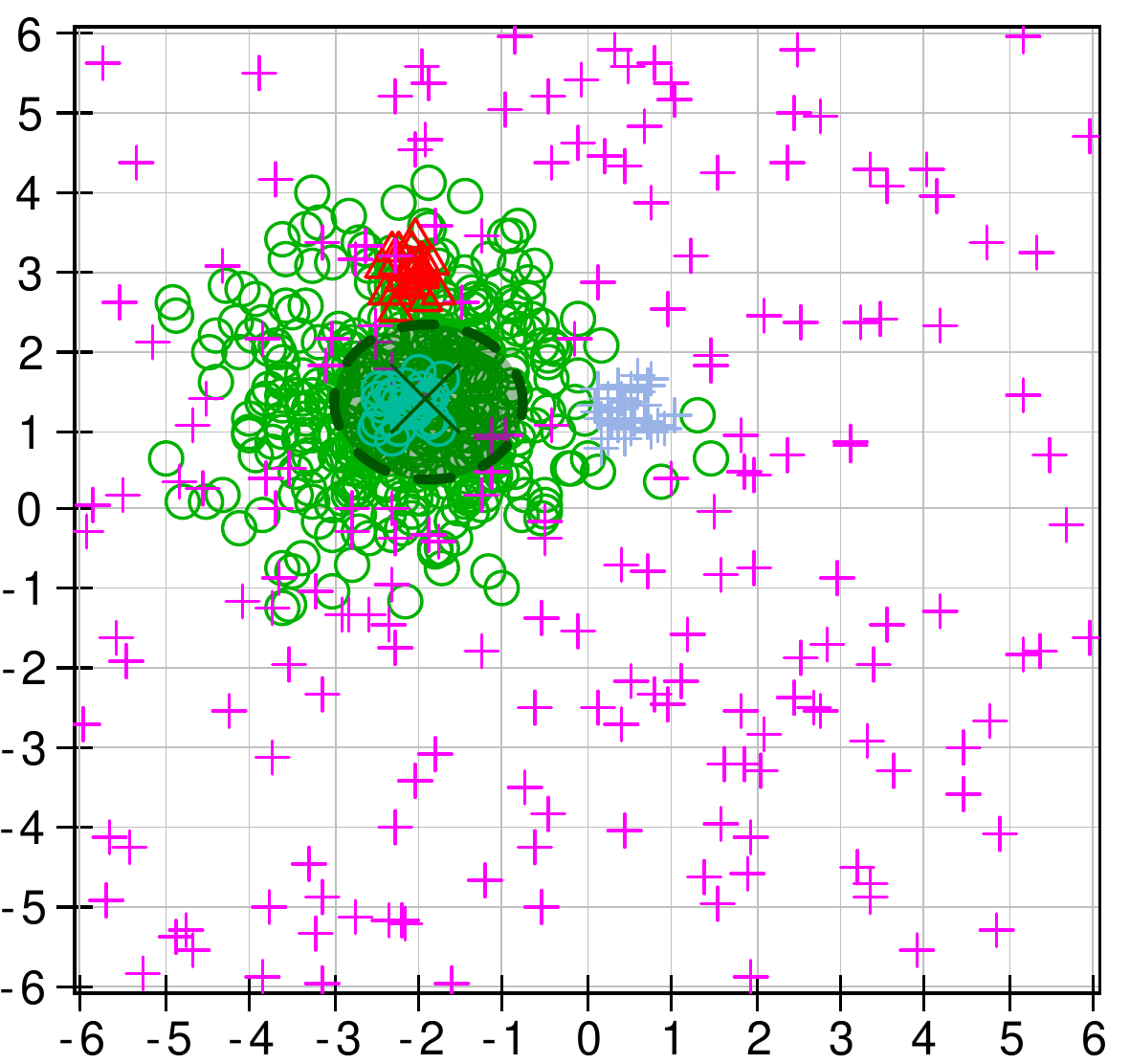}}	
        
        \subfigure[Test statistics over time for non-adjusted test. \label{fig:noise_candies:noest}]{\includegraphics[width=.475\columnwidth]{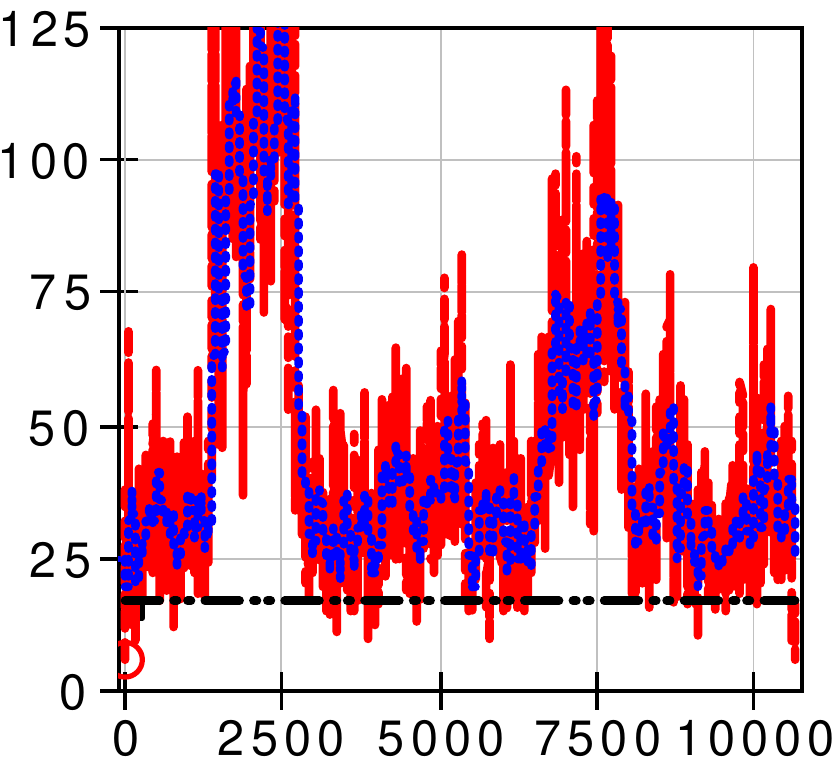}}
        \subfigure[Test statistics over time for estimated distance distribution. \label{fig:noise_candies:est}]{\includegraphics[width=.475\columnwidth]{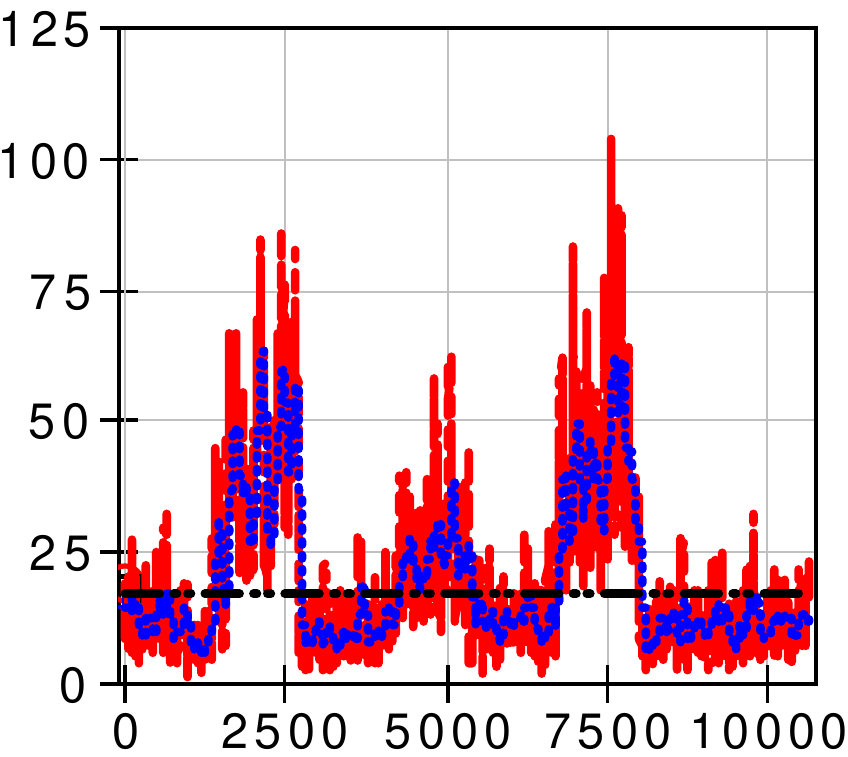}}	
        \caption{Scenario with samples from three \textit{novel} processes and additional uniform distributed noise that is scattered in the input space and corresponding test curves: test statistics (red) and moving average test statistic (blue).\label{fig:noise_candies}}
    \end{center}
    \vskip -0.2in
\end{figure}

Figure \ref{fig:noise_candies:ds} depicts the same exemplary data set that is already used in Section \ref{sec:transform} interspersed with uniform random noise (purple + crosses). The corresponding $t$-value curve is displayed in Figure \ref{fig:noise_candies:noest} and shows a recognizable up-shift, introduced by the noise. Nevertheless, this undesired effect can be circumvented by adjusting the distance distribution according to Section \ref{sec:learned_dist}. The curve of the adjusted test is illustrated in Figure \ref{fig:noise_candies:est}. The course of the \textit{moving-average} is now clearly below the threshold in intervals where no \textit{novelty} is present, but rises clearly - although weaker as compared to the application without noise - above the critical value, when the \textit{novel} processes start to generate samples.
Therefore the high-density approach is essentially capable of handling noise, but requires the presence of noise in the training data.

\subsubsection{An alternative view on \theApproach}

At first the two different approaches to novelty detection for LDR and HDR might seem quite different. It is, however, possible to get a consistent view by interpreting one detector by means of the other. As mentioned before, the last bins $\mathrm{cell}_\lambda$ of each HDR detector matches the low-density parts of the input space. Therefore the ring buffer $\mathcal{B}$ used for 2SND can be seen as a shared \textit{cell} across all HDR detectors. On the other hand, the individual buffers of each HDR detector allow an interpretation as \textit{clusters} (with a different adaptation predicate $P$), and thus suiting the 2nd stage of 2SND.

\section{Case Study}
\label{sec:casestudy}

{\color{blue}
To validate that the presented approach can be used to real-world applications, we show experimental results based on the
well-known \textit{KDD Cup 1999} network intrusion data set \cite{KDD99DS}.
Even though it is pointed out that there are some serious flaws in the data set, which makes it inappropriate for the evaluation of real intrusion detection systems, its properties are still suitable for our purposes, since we are not interested in building a state of the art intrusion detection system.

\subsection{Setup}
}

As mentioned before, our new approach is compared to a novelty detection technique that is proposed in \cite{Fis12}. 
Here, \textit{novel} samples are also identified using a GMM and the squared Mahalanobis distance between processed samples and the mean of the different components.
Each time a new sample is processed, an internal state variable $S_n$ is updated,
such that $S_n = S_{n-1} + \chi^2_{nov}$, with:
\begin{equation}
\label{eq:chi2nov}
\chi^2_{nov}(\mathbf{x}) = \eta \sum_{j=1}^{J}p(j|\mathbf{x}) \left(\delta_{\alpha,j}(\mathbf{x})  - \frac{\alpha}{1-\alpha}(1-\delta_{\alpha,j}(\mathbf{x})) \right) \\
\end{equation}
being a penalty or reward, depending on how well the new sample fits the model. To compute whether the state variable is rewarded or punished, the indicator functions:
\begin{align}
\label{eq:indicator}
\delta_{\alpha,j}(\mathbf{x}) &= \begin{cases}
1, \enskip \Delta^2_j(\mathbf{x}) \leq \rho = F_{{\chi}_{D}^2}^{-1}(\alpha) \\
0, \enskip \text{sonst}
\end{cases}
\end{align}
of each component are evaluated and the results are multiplied with the responsibilities of the components. If the algorithm is processed in an environment without emerging processes, the expectation of the state variable will be equal to its initial value $E[S_n]=1$. The presence of a \textit{novel} process will lead to a decrease of the value of the state variable $S_n$. This can be exploited to detect \textit{novel} processes as soon as the state variable underflows a given threshold (here: $0.2$). The parameter $\eta$ controls how fast the state variable changes (here: $0.001$).
This causes a model adaptation that uses the last $500$ observations to retrain the model, which is done with a modified VI algorithm, that allows to insert new components into an existing GMM and train only those, keeping the existing components ``fixed''.
After the model is adapted to its changed environment, the state variable is reset to its initial value. We refer to this approach as \textbf{CSND} ($\chi^2$-novelty detection) 

{\color{blue}
Originating from the various recorded connections in the KDD99 data set, different attack \textit{scenarios} are sampled (these are: ipsweep, neptune, nmap, satan, and smurf). 
Each scenario consist of background connections (legitimate network traffic) and connections related to the specific attack. A dimension reduction to 6 out of the 41 dimensions is performed as preprocessing step. Additionally and due to the massive support in terms of categories, we interpret the discrete attributes as nearly continuous.
Each scenario consists of three parts with an overall of 25000 connections.
The first part contains 10000 connections drawn from a pool of background connections only.
The second part is a mixture of background and attack connections (with the attack name as label) with a ratio of 3:1 and a total of 10000 connections.
The last 5000 connections form the third part, which again consists only of legitimate traffic.
}

Both \textit{adaptive} classifiers are trained with the first 5000 samples of the first part of each scenario to learn an initial GMM with VI. The experiments themselves are conducted in a 5-fold cross-validation fashion, with independent folds for the \textit{train sets}, which consist of connections from the first and third parts of each scenario, and a single \textit{test set} that is equal to the second part of each scenario.

Additionally, to get a baseline for the classification performance, a static classifier (as described in Section \ref{sec:cmm}, referred to as GMM-Static) is trained on samples of all classes (background connections and attacks). That is, this classifier can be seen as omniscient as it anticipates future attacks that are completely new and unpredictable for the two adaptive classifiers above. In order to get meaningful results, a stratified 5-fold cross-validation, with all connections mixed together, is carried out.
Then the accuracy and the $F_1$-score of the \textit{class} assigned to samples of the \textit{novel process} are used to evaluate the classification performance.

%

\subsubsection{Results}

The resulting averaged classification performances are summarized in Table \ref{tab:kdd:perf}, which states that both adaptive approaches are able to identify the attacks and perform model adaptations that integrate the acquired knowledge. In all scenarios, the accuracy and the observed $F_1$-score of \theApproach~ is equal or higher compared to those of the CSND approach. In three out of five scenarios our approach performs comparably well as the static baseline and still satisfiable on the other two. 
\begin{table}[h]	
    \caption{Comparison of classification accuracies and $F_1$-scores for the \textit{novel process} (in form of the applied attack) of both novelty detection approaches and the GMM-Static baseline. \label{tab:kdd:perf}}
    \vskip 0.15in
    \setlength{\tabcolsep}{3pt} 
    \begin{center}
        \begin{small}
            \begin{sc}
                \begin{tabular}{lcccccc}
                    \hline 
                    \textbf{Scenario} &  \multicolumn{2}{c}{\textbf{\theApproach}}  &\multicolumn{2}{c}{\textbf{CSND}} & \multicolumn{2}{c}{\textbf{GMM-Static}} \\
                    & Acc & $F_{1nov}$ & Acc & $F_{1nov}$ & Acc & $F_{1nov}$ \\ \hline
                    \textbf{ipsweep} & $97.1\%$ & ($0.9$) & $86.1\%$ & ($0.1$) & $96.2\%$ & ($0.9$) \\
                    \textbf{neptune} & $99.3\%$ & ($1.0$) & $94.7\%$ & ($0.8$) & $98.8\%$ & ($1.0$) \\
                    \textbf{nmap}    & $95.3\%$ & ($0.8$) & $90.9\%$ & ($0.4$) & $98.0\%$ & ($0.9$) \\
                    \textbf{satan}   & $95.3\%$ & ($0.7$) & $92.3\%$ & ($0.7$) & $99.0\%$ & ($1.0$) \\
                    \textbf{smurf}   & $99.2\%$ & ($1.0$) & $92.8\%$ & ($0.7$) & $99.8\%$ & ($1.0$) \\
                    & & & & & & \\
                    $\o$ & $97.2\%$ & ($0.9$) & $91.4\%$ & ($0.5$) & $98.4\%$ & ($0.9$) \\
                    \hline 
                \end{tabular}
            \end{sc}
        \end{small}
    \end{center}
    \vskip -0.1in	
\end{table}

The higher performance of \theApproach~ over CSND is explained by Table \ref{tab:kdd:seen}, which shows the average number of \textit{actual novel} samples (samples actually belonging to the attack) that are processed before the \textit{novel process} is detected and a model adaptation triggered.
\begin{table}[h]
    \caption{Number of actual \textit{novel} samples needed until a \textit{novel} process gets detected and a model adaptation is triggered.\label{tab:kdd:seen}}
    \vskip 0.15in
    \begin{center}
        \begin{small}
            \begin{sc}
                \begin{tabular}{lcc}
                    \hline 
                    \textbf{Scenario} & \textbf{\theApproach} & \textbf{CSND} \\ 
                    & \multicolumn{2}{c}{required observations} \\
                    \hline
                    \textbf{ipsweep} & $40.4$ & $1063.4$ \\
                    \textbf{neptune} & $50.6$ & $1043.0$ \\
                    \textbf{nmap}    & $85.0$ & $1173.2$ \\
                    \textbf{satan}   & $19.4$ & $1151.8$ \\
                    \textbf{smurf}   & $37.2$ & $1041.4$ \\
                    & & \\
                    $\o$ & $46.5$ & $1094.6$ \\
                    \hline 
                \end{tabular}
            \end{sc}
        \end{small}
    \end{center}
    \vskip -0.1in	
\end{table}
Here \theApproach~ displays its strength to exploit spatial information between \textit{suspicious} samples in LDR the form of clusters, which accelerates the detection compared to the slowly changing state variable of CSND.

The algorithm is designed to be processed in an online mode. Therefore, the number of triggered model adaptation steps and the number of inserted components are also investigated.
Table~\ref{tab:kdd:adapt} shows the averaged number of adaptation and insertion steps for each scenario.
\begin{table}[h]
    \caption{Number of triggered model adaptations and average number of inserted components (in parentheses) for both approaches. \label{tab:kdd:adapt}}
    \vskip 0.15in
    \begin{center}
        \begin{small}
            \begin{sc}
                \begin{tabular}{lcccc}
                    \hline
                    \textbf{Scenario} &  \multicolumn{2}{c}{\textbf{\theApproach}}  &\multicolumn{2}{c}{\textbf{CSND}} \\
                    & Adapt. & Comp. & Adapt. & Comp. \\ \hline
                    \textbf{ipsweep} & $1.0$ & ($3.0$) & $2.0$ & ($13.6$) \\
                    \textbf{neptune} & $1.0$ & ($1.8$) & $1.0$ & ($1.0$) \\
                    \textbf{nmap}    & $1.2$ & ($3.2$) & $1.0$ & ($5.8$) \\
                    \textbf{satan}   & $2.0$ & ($5.0$) & $1.0$ & ($4.8$) \\
                    \textbf{smurf}   & $1.6$ & ($3.0$) & $1.0$ & ($7.8$) \\
                    & & & & \\
                    $\o$ & $1.4$ & ($3.2$) & $1.2$ & ($6.6$) \\
                    \hline 
                \end{tabular}
            \end{sc}
        \end{small}
    \end{center}
    \vskip -0.1in	
\end{table}
As we can see, both approaches tend only to a single model adaptation, which is the optimum here. 
The CSND approach has fewer model adaptations on average than the \theApproach, but has a higher average number of inserted components, which is not negligible since the number of components in the GMM has a direct influence on the run-time of both algorithms.

\section{Conclusion and Outlook}
\label{sec:outlook}

We introduced \theApproach, a holistic approach to novelty detection for (new) emerging processes throughout the complete input space of a probabilistic classifier.
To achieve this, different novelty detectors for \textit{low-density} regions (LDR, where it is less likely to observe samples) and \textit{high-density} regions (HDR, samples are expected to be observed here) \comment{Nochaml abkürzungen einführen, oder ganz weglassen?} are combined
and thus able to cover the complete input space.  \comment{Impliziert, dass der gesamte Eingaberaum aus eben diesen (LDR,HDR) besteht.}
For LDR we resort on 2SND, this algorithm works with two stages. First, \textit{suspicious} observations are identified with the help of a GMM (which are based on parametric densities). In the second stage, \textit{suspicious} samples are then clustered in a nonparametric way (inspired by DBSCAN). A \textit{novel} process is recognized as soon as one of the (nonparametric) clusters reaches a sufficient size.
The detection in HDR on the other hand is purely based on \textit{parametric density estimation}. We showed how to use multiple detectors (one detector per component) to identify \textit{novelty} in GMM. The presence of \textit{novel processes} in HDR is directly identified. This is accomplished by maintaining a \textit{sliding window} of recent observations and performing statistical \textit{goodness-of-fit} tests between \textit{sliding window} and the affiliated component.

In a compact case study in the field of computer network intrusion detection, we could show that \theApproach~ is applicable to real-world data sets. We tested it on a subset of the well-known KDD Cup '99 Intrusion Detection data set, where rather promising results were obtained. So far, first experiments on artificial laboratory data sets lead us to the conclusion that \theApproach~ will be a satisfactory solution to \textit{novelty detection} with model adaptation in the near future. \comment{hier kam noch die Modelladaption hinzu}

In our future work we will focus on extending the described \textit{novelty} detector further, this includes in particular \textit{reaction} procedures for the HDR detection.
We will elaborate the performance of \theApproach~ on more sample applications, e.g., in the fields of robotics or video based surveillance. Detection and handling of obsoleteness or concept shift will be accomplished with techniques similar to the ones proposed here. The same holds for concept drift, but here, it will be quite difficult to effect the trade-off between under- and overreaction (too early or too late). The accuracy of our techniques must be set in relation to a ``degree'' of time-variance in the observed system. 
It will be possible to detect emergent phenomena in the observed environment and to numerically assess the degree of emergence (cf.\ \cite{Fis10}). Also, these techniques allow for an application to various anomaly detection problems.
Furthermore, the design of the approach is not necessarily limited to GMM but applicable to other mixture models as well.

\comment{hier noch die awareness unterbringen}
{\color{red}
Another possible application field that could benefit from our proposed technique are systems equipped with \textit{awareness} capabilities. Often, terms such as location-aware, context-aware, self-aware, or en\-vironment-aware are used in the literature (see, e.g., \cite{Abowd99,OC-Book}). In our opinion, awareness is essentially 
\begin{itemize}
    \item the capability to compare knowledge about the self, the environment, other systems etc.\ to current observations in order to detect when expectations concerning current observations do not meet the actual observations anymore and
    \item the ability to adapt the knowledge model in a way such that the system meets some performance requirements which includes a solution to the problem when to adapt the model in odrder to avoid a performance loss either due to too fast or too slow reactions.
\end{itemize}
Altogether, awareness techniques will be a key to develop new kinds of technical systems that could actually be termed to be ``intelligent'' or ``smart'' with some higher degree of justification.}

\section*{Acknowledgments}
The authors would like to thank the German Research
Foundation (DFG) for support within the DFG project
CYPHOC (SI 674/9-1).

\bibliography{./lit}
\bibliographystyle{abbrv}

\end{document}